\documentclass{article}

\usepackage[preprint]{neurips_2025}

\usepackage[utf8]{inputenc} 
\usepackage[T1]{fontenc}    
\usepackage{hyperref}       
\usepackage{url}            
\usepackage{booktabs}       
\usepackage{amsfonts}       
\usepackage{nicefrac}       
\usepackage{microtype}      
\usepackage{xcolor}         
\usepackage{makecell}
\usepackage{microtype}
\usepackage{hyperref}
\usepackage{url}
\usepackage{verbatim}
\usepackage{listings}
\usepackage{booktabs}
\usepackage{graphicx}
\usepackage{lineno}
\usepackage{subcaption}
 \usepackage{multirow} 
\usepackage{amsmath} 
\usepackage[most]{tcolorbox}
\usepackage{fontawesome5} 
\usepackage{xcolor}
\usepackage{amssymb}

\definecolor{promptblue}{RGB}{41, 98, 255}
\definecolor{lightblue}{RGB}{240, 247, 255}
\definecolor{codegray}{RGB}{245, 245, 245}
\definecolor{darkblue}{rgb}{0, 0, 0.5}
\title{Between Underthinking and Overthinking: An Empirical Study of Reasoning Length and correctness in LLMs}

\author{Jinyan Su$^1$, Jennifer Healey$^2$,
  Preslav Nakov$^3$,  Claire Cardie$^1$\\
  $^1$ Cornell University, 
  $^2$ Adobe Research,
  $^3$ MBZUAI\\
  \texttt{\{js3673, ctc9\}@cornell.edu, jehealey@adobe.com, preslav.nakov@mbzuai.ac.ae}
  }

\begin{document}

\maketitle

\begin{abstract}
Large language models (LLMs) are increasingly optimized for long reasoning, under the assumption that more reasoning leads to better performance. However, emerging evidence suggests that longer responses can sometimes degrade accuracy rather than improve it. In this paper, we conduct a systematic empirical study of the relationship between reasoning length and answer correctness. We find that LLMs tend to overthink simple problems—generating unnecessarily long outputs—and underthink harder ones, failing to extend their reasoning when it is most needed, indicating that models might misjudge problem difficulty and fail to calibrate their response length appropriately. Furthermore, we investigate the effects of length reduction with preference optimization algorithm when simply preferring the shorter responses regardless of answer correctness. Experiments show that the generation length can be significantly reduced while maintaining an acceptable accuracy. Our findings highlight generation length as a meaningful signal for reasoning behavior and motivate further exploration into LLM's self-awareness in reasoning length adaptation.
\end{abstract}

\section{Introduction}
Recently, the emergence of reasoning large language models (LLMs), such as OpenAI O1 \citep{openAI-o1} and DeepSeek-R1 \citep{guo2025deepseek}, has demonstrated impressive System-2 thinking capabilities \cite{li2025system}. These models can perform deliberate, multi-step reasoning, which has led to significant improvements in tasks involving logic, mathematics, and complex inference. As a result, there is growing interest in promoting long chain-of-thought (CoT) reasoning, which enables broader exploration and more structured problem-solving \cite{chen2025towards}. Follow-up models \citep{QwQ, Flash-thinking} and recent research \citep{chen2025towards, muennighoff2025s1} further reinforce the trend of scaling up reasoning length to enhance performance.

However, longer reasoning is not always better. In some cases, shorter answers are not only sufficient but actually more accurate, whereas longer chains introduce unnecessary steps or compounding errors. Recent studies have shown that accuracy often plateaus—and may even decline—once reasoning length exceeds a certain threshold \citep{xie2025logic, jin2024impact, wu2025more}. As \citet{yang2025towards} point out, the excessive pursuit of lengthy reasoning can negatively impact a model’s performance. This overthinking phenomenon has been further pointed out in works such as \citet{chen2024not} and \citet{ luo2025o1}, which highlight how current reasoning LLMs tend to generate overly verbose outputs even for simple problems, often with minimal or no improvement in accuracy.

While emerging evidence suggests that the relationship between reasoning length and correctness is more complex than previously assumed \citep{xie2025logic, jin2024impact, wu2025more}, it has not yet to be systematically studied. In particular, it remains unclear when longer reasoning improves accuracy, and when it simply adds computational and latency overhead without any accuracy gain. A more comprehensive investigation into these dynamics is essential for building models that reason both effectively and adaptively.

To address this gap, we conduct a systematic empirical analysis of the relationship between reasoning length and answer correctness. Our study uses two popular reasoning-capable models—DeepSeek-1.5B-Distill \citep{deepseekai2025deepseekr1incentivizingreasoningcapability} and DeepScaler-1.5B-Preview \citep{deepscaler2025}—evaluated on two widely used math reasoning benchmarks: GSM8K \citep{cobbe2021gsm8k} and MATH \citep{hendrycks2021measuring}. We analyze the relationship between generation length and accuracy from both the sample level and the question level.
 In sample level, we study how generation length influence the correctness for a fixed question. At the question level, we examine whether models adapt their generation length according to perceived question difficulty, and if they accurately recognize and respond to this difficulty.

Through empirical analysis, we aim to understand how generation length affects answer correctness, investigate patterns of model behavior related to overthinking and underthinking, and examine how these patterns correspond to the model's perceived difficulty of the problem. Our findings are:
\begin{itemize} 
\item For a fixed question, accuracy exhibits a non-monotonic relationship with reasoning length: it tends to increase with longer responses up to a point, but declines when the reasoning becomes excessively long.
\item Questions that are answered incorrectly tend to have longer average reasoning lengths than correctly answered ones, likely due to their inherent complexity and the need for more reasoning steps. However, models do not consistently recognize or adapt to problem difficulty. For relatively easy questions, models are often able to detect smaller increases in difficulty and appropriately extend their reasoning. In contrast, when encountering problems beyond their capabilities, models sometimes \textbf{underthink}—either failing to recognize the increased difficulty or lacking the necessary knowledge to respond effectively. As a result, they often generate responses that are shorter than needed to arrive at a correct answer.
\end{itemize}
Building on our analysis, we further investigate the effect of length based preference optimization. Specifically, we explore whether encouraging the model to produce shorter responses—without directly optimizing for correctness—can still preserve answer accuracy. Using only unlabeled data, we fine-tune the model to prefer shorter responses and find that this preference tuning can maintain relatively strong accuracy while reducing token length, even in the absence of ground-truth supervision. Though the token length reduction is largely due to the reduced length for the incorrect responses, since they are significantly longer than correct responses and easier to reduce. However, we also observe a 10\%–25\% reduction in the length of correct responses. This suggests that while the model is capable of recognizing question difficulty and adjusting its generation length accordingly for easier problems, a tendency toward \textbf{overthinking} still exists.
Our findings and experiments motivate future work on enhancing LLMs' self-awareness in reasoning length adaptation.

\section{Related Work}
\paragraph{Concise thinking} \cite{team2025kimi} observes the issue of overthinking, i.e., excessively lengthy reasoning during RL training and uses a length reward to restrain the increase in response token length. \cite{chen2024not} provides a comprehensive analysis for overthinking and explores several mitigation.  \cite{munkhbat2025self} proposes to use self-generated concise reasoning paths obtained by best-of-N sampling to elicit LLM's latent ability for concise generation. \cite{luo2025o1} proposes O1-Pruner, first estimates the LLM’s baseline performance through pre-sampling and then uses RL-style fine-tuning to encourage the model to
generate shorter reasoning processes under accuracy constraints.  
These above efficient thinking works mainly focuses on reducing the thinking cost while maintain a relatively good accuracy. \cite{fu2025reasoning} proposes Dynasor-CoT, a certainty-based approach for dynamic reasoning termination.  \cite{yeo2025demystifying, zhang2025grpo, aggarwal2025l1} modifies the reward to reduce the generation length. \cite{ma2025reasoning} found that  simply prompt the LLM to skip the thinking process can  achieve competitive performance, encouraging a reconsideration of the necessity of
lengthy thinking processes.
\paragraph{Adaptive Thinking}
\cite{han2024token} proposes a token budget aware LLM reasoning framework that
dynamically adjusts the number of reasoning tokens based on the reasoning complexity of each
problem. \cite{xu2024adaption} adaptively
adjust the complexity of the prompting based on the difficulty of the questions; \cite{damani2024learning, wang2024make, manvi2024adaptive, li2024escape} focuses on inference time compute cost, either adaptively allocate test time compute or use early stopping to save the cost while \cite{shen2025dast}  enables models to autonomously adjust the length of CoT based on problem difficulty. \cite{pan2024dynathink} dynamically switch between fast and slow thinking based on model confidence. 
\cite{ma2025cot}
enables elastic control of length for CoT
within the parameter space, allowing a single model
to generate CoT from short to long. \cite{yang2025towards} found that there exists an optimal scaled length distribution that differs across different domains. \cite{ballon2025relationship} shows it is not necessary to generate longer chain of thought in order to achieve high accuracy. \cite{qu2025optimizing} learn a strategy that is agnostic of the test-time budget, which, when deployed, the LLM spends only the necessary amount of
tokens while still making progress when run at larger token budgets.
\paragraph{Optimal Thinking}
\cite{wu2025more}  observes that as the number of reasoning steps increases, performance initially improves but eventually decreases, theoretically prove the existence of an optimal CoT length, highlighting the critical need to calibrate CoT length to align with model capabilities and task demands. 
\section{Experimental Setting}
\paragraph{Models and Dataset}
Our study mainly uses two reasoning models—DeepSeek-1.5B-Distill \citep{deepseekai2025deepseekr1incentivizingreasoningcapability} (Denoted as \texttt{R1-Distill}) and DeepScaler-1.5B-Preview (Denoted as \texttt{R1-Preview})\citep{deepscaler2025}—evaluated on two widely used math reasoning benchmarks: GSM8K \citep{cobbe2021gsm8k} and MATH \citep{hendrycks2021measuring}. For each question  $q\in D$, we generate $N=10$ samples diverse reasoning paths  using temperature $T=1.0$, top-$p=1$ and setting max token length as 8k:
 \begin{equation*}
 \{(o_i^{(q)}, l_i^{(q)}, c_i^{(q)})\}_{i=0}^{N-1}
 \end{equation*}
 Here $o_i^{(q)}$ denotes the output of the $i$-th sample, $l_i^{(q)}$ is the number of output tokens (i.e., the reasoning length), and $c_i^{(q)}\in \{0,1\}$ indicates whether the answer is correct  according to the ground truth.

\section{Sample-Level Analysis}

In this section, we examine how reasoning length influences answer correctness by analyzing multiple generated samples for the same input. This sample-level analysis isolates variation in reasoning length that arises purely from the model’s own generation process, rather than the problem difficulty. Formally, we fix a question and a model, and allow the model to generate multiple responses via sampling without external intervention. We then ask: what is the relationship between the reasoning length of each sample and its correctness?
Unlike question-level analysis—which intertwines reasoning length and correctness with the underlying difficulty of each question—sample-level analysis allows us to more precisely assess the impact of generation length on correctness, while holding all other factors to be constant.

\subsection{Non-Linear Relationship of Sample Length and Correctness}
\begin{figure}[h]
    \centering
    \includegraphics[width=1\linewidth]{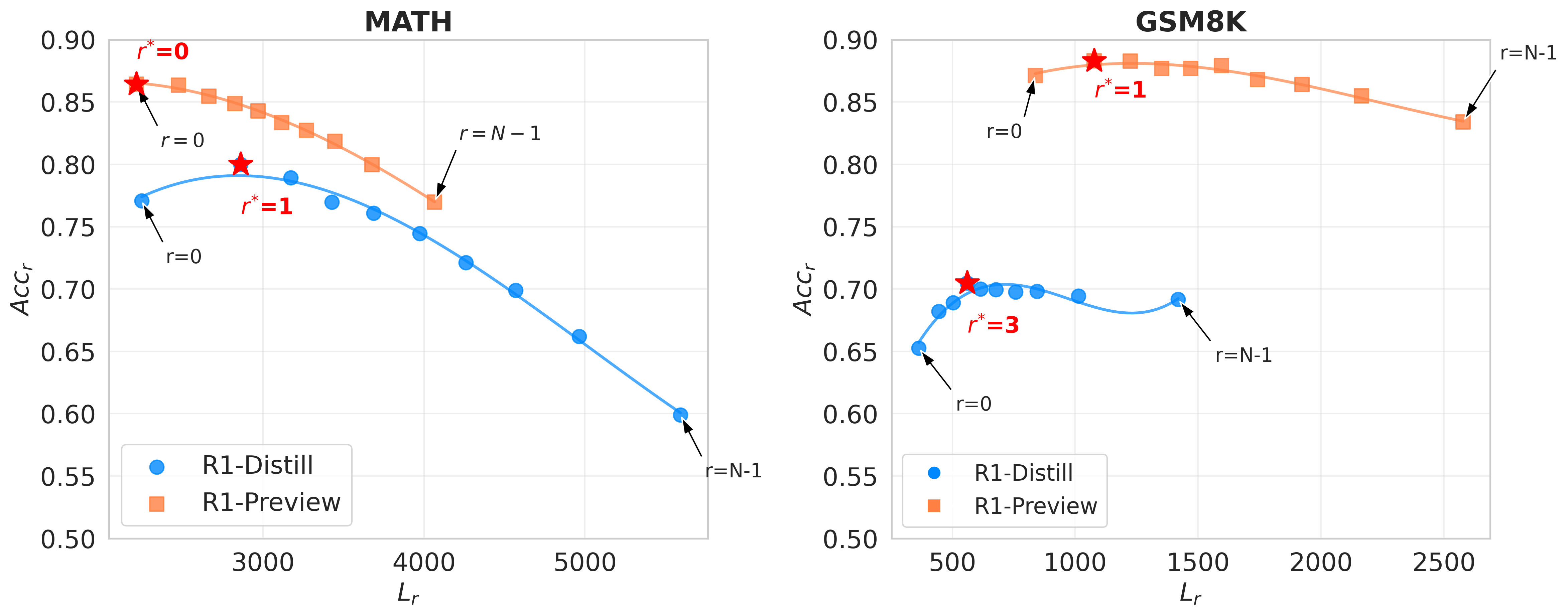}
    \caption{Reasoning length $L_r$ of the $r$-th shortest response v.s. accuracy $\text{Acc}_r$ ($N=10$).  The red marker \textcolor{red}{$r^*$} denotes the rank of responses with the highest accuracy. Results for R1-Distill and R1-Preview on the GSM8K and MATH datasets suggest that both overly short and excessively long reasoning can degrade performance.}
    \label{fig: token len vs acc}
\end{figure}

\paragraph{Accuracy Across Length-Ranked Samples}
Figure~\ref{fig: token len vs acc} shows the relationship between average reasoning length and accuracy across samples ranked by their length. Here, $L_r$ and $\text{Acc}_r$ denote the average reasoning length and accuracy of the $r$-th shortest sample, respectively. For example, $L_0$ is the average length of the shortest response for each question among all the responses for that question. As the rank $r$ increases (corresponding to longer reasoning on average), we observe a consistent non-monotonic trend across models and datasets. For the R1-Distill model, accuracy initially improves with increasing reasoning length, but beyond a certain point, further increases in length lead to a decline in performance. The peak accuracy is achieved at $r^{*}=1$ on the MATH dataset and at $r^{*}=3$ on GSM8K. 
The R1-Preview model follows a similar trend on GSM8K, with the highest accuracy at  $r^{*}=1$. However, on the MATH dataset, R1-Preview achieves its best performance at $r^{*}=0$, with accuracy declining at higher ranks—suggesting that for this model-dataset pair, concise reasoning is not only sufficient but optimal.
Overall, these patterns reveal a non-linear relationship between reasoning length and correctness: while overly brief responses often lack sufficient reasoning to produce correct answers, excessively long outputs tend to be detrimental to both accuracy and efficiency. These results point to the existence of an optimal reasoning length within the sample distribution—neither too short nor too long—for achieving the best accuracy while maintaining efficiency.
\paragraph{Shortest Correct Response}
Figure~\ref{fig: first correct index} shows the percentage of questions for which the shortest correct response appears at rank $i$. We find that for over 60\% of questions—across all model and dataset combinations—the shortest sampled response is already correct. Among questions that the model is capable of answering (i.e., those with at least one correct response), the correct answer is often found within the first few shortest responses. Additionally, we analyze other rank related statistics, including the rank of the longest correct response, the shortest and longest incorrect responses, and the correlation between reasoning length and correctness. Due to space constraints, we refer interested readers to Appendix~\ref{app: sample level additional analysis} for these extended results.
\begin{figure}[h]
    \centering
    \includegraphics[width=1\linewidth]{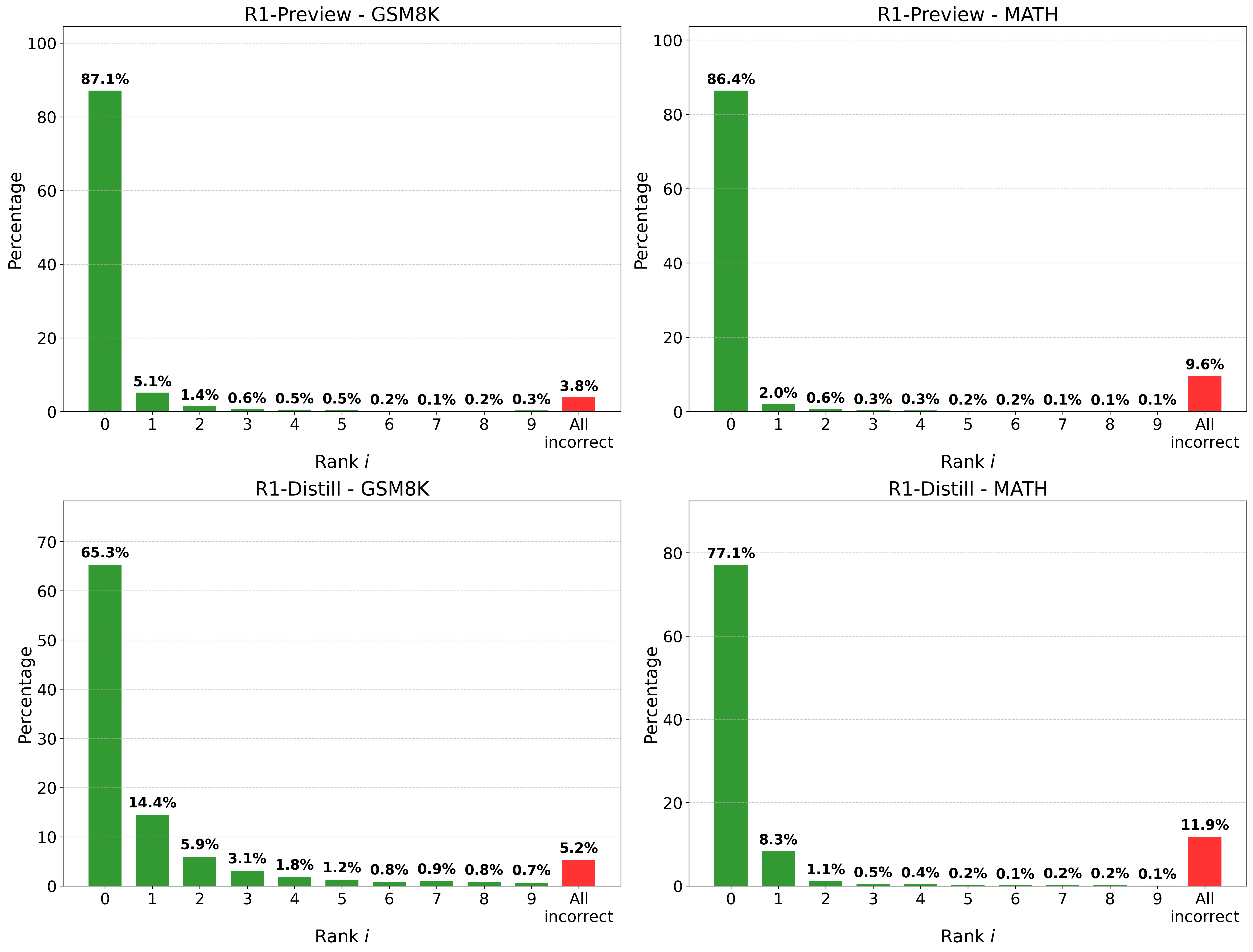}
    \caption{Percentage of questions for which the shortest correct response occurs at rank $i$.}
    \label{fig: first correct index}
\end{figure}

The findings from our sample-level analysis suggest that correct responses are often concentrated among the shorter or moderately long samples—where "short" and "moderate" are defined relative to the overall distribution of response lengths for a fixed question and model. We conjecture that this pattern arises because, once the $i$-th shortest response is already correct, generating longer responses beyond that point amounts to \textbf{overthinking}, i.e., it increases the risk of introducing errors or irrelevant information and compounding small mistakes, which can ultimately derail the reasoning process and reduce the likelihood of producing a correct answer.
\section{Question-Level Analysis}
At the question level, the relationship between response length and answer correctness becomes even more pronounced.  We begin by comparing the average lengths of correct and incorrect responses. As shown in Table~\ref{tab:token_length_correlations}, incorrect responses are significantly longer than correct ones across all four model–dataset combinations. This trend is especially striking on the MATH dataset, where the average length of incorrect responses exceeds 6,000 tokens, while correct responses average fewer than 3,000 tokens for both models.  In addition, we observe consistently strong negative Pearson and Spearman correlations between response length and accuracy—particularly on the MATH dataset—suggesting that longer responses are generally associated with lower correctness.
\begin{table}[ht]
\centering
\caption{Correlation between token length and accuracy  across four model-dataset combinations. We also report the average token length for correct responses ($L^{\checkmark}$) and incorrect responses ($L^{\times}$).}
\label{tab:token_length_correlations}
\scriptsize
\begin{tabular}{llcccc}
\toprule
\textbf{Model} & \textbf{Dataset}&\textbf{\makecell{Pearson \\ Correlation}} & \textbf{\makecell{Spearman \\ Correlation}  } & \textbf{\makecell{ $L^{\checkmark}$ (Correct)\\ (Mean ± Std)} } & \textbf{\makecell{ $L^{\times}$ (Incorrect)\\  (Mean ± Std)} } \\
\midrule
\multirow{2}{*}{R1-Preview } & GSM8K & -0.5347 & -0.3740 & 1374.82 $\pm$ 92.34 & 3069.22 $\pm$ 337.53 \\
& MATH & -0.6835 & -0.6390 & 2459.84 $\pm$ 29.92 & 6106.17 $\pm$ 37.70 \\\midrule
\multirow{2}{*}{R1-Distill }&GSM8K & -0.1626 & -0.2216 & 666.68 $\pm$ 131.59 & 842.46 $\pm$ 187.08 \\
& MATH & -0.7248 & -0.6801 & 2907.23 $\pm$ 275.25 & 6521.31 $\pm$ 433.08 \\
\bottomrule
\end{tabular}
\end{table}

\subsection{Impact of Question Difficulty on Response Length}
To better understand why incorrect responses are significantly longer in the question level,  we categorize questions based on model-specific difficulty levels. Specifically, we define three difficulty tiers: \textit{Easy}, \textit{Medium}, and \textit{Hard}. A question is labeled \textit{Easy} if the model answers it correctly in all $N=10$ sampled responses, \textit{Hard} if it fails to answer it correctly in all 10 responses, and \textit{Medium} if correctness varies across sampled responses. The number of questions in each category and their corresponding average response lengths are shown in Table~\ref{tab:question-distribution} and Figure~\ref{fig: token length by difficulty} respectively (see Appendix~\ref{app: statistics for each difficulty level}). Consistent with the trends observed in Table~\ref{tab:token_length_correlations}, we find that \textit{Easy} questions—those consistently answered correctly by the model—tend to have the shortest responses, while \textit{Hard} questions—those consistently answered incorrectly—produce the longest. However, it remains unclear whether the increased length reflects the model’s ability to recognize the intrinsic complexity of the question and adapt its reasoning accordingly, or whether the longer responses themselves introduce additional errors, making the question appear more difficult than it truly is.
To disentangle these factors, we start by a cross-model analysis that examines response lengths within the subset of questions labeled as \textit{Easy} for a given model.

Specifically, for two models $\mathcal{M}_i$ and $\mathcal{M}_j$, we compare generation lengths on three sets of questions: (1) the shared-easy set $Q_{\cap}^{\text{easy}} = Q_i^{\text{easy}} \cap Q_j^{\text{easy}}$, which both models find easy; (2) the $\mathcal{M}_i$-advantage set, $Q_{i \backslash j}^{\text{easy}} = Q_i^{\text{easy}} \setminus Q_j^{\text{easy}}$, which is easy for $\mathcal{M}_i$ but not for $\mathcal{M}_j$; and (3) the $\mathcal{M}_j$-advantage set, $Q_{j \backslash i}^{\text{easy}} = Q_j^{\text{easy}} \setminus Q_i^{\text{easy}}$, which is easy for $\mathcal{M}_j$ but not for $\mathcal{M}_i$. If model $\mathcal{M}_i$ generates longer responses on the $\mathcal{M}_i$-advantage set than on the shared-easy set, it suggests that questions in the $\mathcal{M}_i$-advantage set are intrinsically harder for $\mathcal{M}_i$ than those in the shared-easy set—potentially requiring longer reasoning. A similar interpretation holds for model $\mathcal{M}_j$. 
\begin{table}[h]
\centering
\caption{T-test results comparing response lengths on  the shared-easy set $Q_{\cap}^{\text{easy}}$ with $\mathcal{M}_1$-advantage set $Q_1^{\text{easy}}\backslash Q_2^{\text{easy}}$ and $\mathcal{M}_2$-advantage set $Q_2^{\text{easy}}\backslash Q_1^{\text{easy}}$ on $\mathcal{M}_1$ (R1-Preview) and $\mathcal{M}_2$ (R1-Distill), respectively. Significance is determined at $p < 0.05$.}
\label{tab: t test for easy questions}
\small
\begin{tabular}{lc|cc|cc}
\toprule
\multicolumn{2}{c|}{}  
& \multicolumn{2}{c|}{\textbf{GSM8K}} 
& \multicolumn{2}{c}{\textbf{MATH}} \\
\midrule
\multirow{6}{*}{\makecell{$\mathcal{M}_1$\\(R1-Preview)}} &
& \textbf{$\mathcal{M}_1$-Adv Set}
& \textbf{$\mathcal{M}_2$-Adv Set}
& \textbf{$\mathcal{M}_1$-Adv Set}
& \textbf{$\mathcal{M}_2$-Adv Set} \\\cline{2-6}
&\textbf{Avg. Tokens on $Q_{\cap}^{\text{easy}}$}  &\multicolumn{2}{c|}{1015.37} &\multicolumn{2}{c}{1690.62}\\\cline{2-6}
& \textbf{Avg. Tokens $Q^{\text{easy}}_{i \backslash j} $}& 1348.60 & 1280.41 
& 2832.86 & 2961.75 \\\cline{2-6}
&\textbf{T-Statistic} & 17.7067 & 7.1635 
& 41.5385 & 14.7609 \\
& \textbf{P-Value} & $5.16\times 10^{-68}$ & $1.08\times 10^{-12}$ 
& $0.00\times 10^{+00}$ & $7.09\times 10^{-48}$ \\\cline{2-6}
& \textbf{Significant?}& Yes & Yes 
& Yes & Yes \\
\midrule
\midrule
\multirow{5}{*}{\makecell{$\mathcal{M}_2$\\(R1-Distill)}} &\textbf{Avg. Tokens on $Q_{\cap}^{\text{easy}}$} &\multicolumn{2}{c}{596.78} &\multicolumn{2}{c}{2322.63}\\\cline{2-6}
& \textbf{Avg. Tokens $Q^{\text{easy}}_{i \backslash j} $}& 632.46 & 555.54 
& 4012.31 & 3126.20 \\\cline{2-6}
& \textbf{T-Statistic}&2.1656 & -1.3379 
& 37.9297 & 6.2503 \\
& \textbf{P-Value} & $3.04\times 10^{-2}$ & $1.81\times 10^{-1}$ 
& $4.87\times 10^{-279}$ & $4.59\times 10^{-10}$ \\\cline{2-6}
& \textbf{Significant?}& Yes & No 
& Yes & Yes \\
\bottomrule
\end{tabular}
\end{table}

\paragraph{Length Signals Perceived Difficulty in Easy Questions} Table~\ref{tab: t test for easy questions} reports the average token lengths and t-test statistics comparing these sets. We observe that for both models, the average response length increases on their respective advantage sets compared to the shared-easy set. For R1-Preview, the average token length increases from 1015.37 on the shared-easy set to 1348.60 on its own advantage set for GSM8K, and from 1690.62 to 2832.86 for MATH.
This suggests that although R1-Preview consistently answers questions in its advantage set correctly, these questions may inherently require longer responses than those in the shared-easy set. Similarly, for R1-Distill, the average response length on MATH increases from 2,322.63 to 3,126.20 when comparing shared-easy to advantage-set questions. In contrast, while R1-Distill’s response length decreases from 596.78 to 555.54 on its advantage set for GSM8K, this reduction is not statistically significant based on the $p$-value. Overall, these patterns indicate that both models generally treat their advantage-set questions as more difficult than those in the shared-easy set, as reflected in the increased length of their responses. Furthermore, we observe that both R1-Distill and R1-Preview's generation length has increased in another model's advantage set. This suggests that even when a model is unable to consistently answer such questions correctly, it may still recognize their increased difficulty and attempt to adapt by generating longer responses.
\begin{table}[h]
\centering
\caption{T-test results comparing response lengths on  shared-sard set $Q_{\cap}^{\text{hard}}$ with  $\mathcal{M}_2$-Advantage set $Q_1^{\text{hard}}\backslash Q_2^{\text{hard}}$ and $\mathcal{M}_1$-Advantage set $Q_2^{\text{hard}}\backslash Q_1^{\text{hard}}$ on $\mathcal{M}_1$ (R1-Preview) and $\mathcal{M}_2$ (R1-Distill), respectively. Significance is determined at $p < 0.05$.}
\label{tab: t test for hard questions}
\small
\begin{tabular}{lc|cc|cc}
\toprule
\multicolumn{2}{c|}{}  
& \multicolumn{2}{c|}{\textbf{GSM8K}} 
& \multicolumn{2}{c}{\textbf{MATH}} \\
\midrule
\multirow{6}{*}{\makecell{$\mathcal{M}_1$\\(R1-preview)}} &
& \textbf{$\mathcal{M}_2$-Adv Set}
& \textbf{$\mathcal{M}_1$-Adv Set}
& \textbf{$\mathcal{M}_2$-Adv Set}
& \textbf{$\mathcal{M}_1$-Adv Set} \\\cline{2-6}
&\textbf{Avg. Tokens on $Q_{\cap}^{\text{hard}}$}  &\multicolumn{2}{c|}{2751.74} &\multicolumn{2}{c}{5845.98}\\\cline{2-6}
& \textbf{Avg. Tokens $Q^{\text{hard}}_{i \backslash j}$}& 3826.86 & 3076.68 
& 5240.06 & 6646.30 \\\cline{2-6}
&\textbf{T-Statistic} & 4.3064 & 1.6509 
& -2.2531 & 5.1719 \\
& \textbf{P-Value} & $2.29\times 10^{-5}$ & $9.96\times 10^{-2}$ 
& $2.46\times 10^{-2}$ & $2.87\times 10^{-7}$ \\\cline{2-6}
& \textbf{Significant?}& Yes & No 
& Yes & Yes \\
\midrule
\midrule
\multirow{6}{*}{\makecell{$\mathcal{M}_2$\\(R1-Distill)}} &\textbf{Avg. Tokens on $Q_{\cap}^{\text{hard}}$} &\multicolumn{2}{c}{1253.09} &\multicolumn{2}{c}{6499.55}\\\cline{2-6}
& \textbf{Avg. Tokens $Q^{\text{hard}}_{i \backslash j}$}& 1492.40 & 1001.46 
& 5411.75 & 7386.94 \\ \cline{2-6}
& \textbf{T-Statistic}& 1.2341 & -1.7563 
& -4.3546 & 6.1999 \\
& \textbf{P-Value} & $2.18\times 10^{-1}$ & $7.98\times 10^{-2}$ 
& $1.53\times 10^{-5}$ & $8.65\times 10^{-10}$ \\\cline{2-6}
& \textbf{Significant?}& No & No 
& Yes & Yes \\
\bottomrule
\end{tabular}
\end{table}

\paragraph{Models Do Not Consistently Recognize or Reflect the Increased Difficulty in Hard Questions} In Table \ref{tab: t test for hard questions}, we analyze how response lengths vary across three subsets of hard questions: the Shared-Hard set $Q^{\text{hard}}_{\cap}$ (questions that are hard by both models), the $\mathcal{M}_2$-advantage set $Q_1^{\text{hard}} \setminus Q_2^{\text{hard}}$ (questions that are hard for $\mathcal{M}_1$ but of lower difficulty for  $\mathcal{M}_2$), and the $\mathcal{M}_1$-advantage set $Q_2^{\text{hard}} \setminus Q_1^{\text{hard}}$. Following the logic from our analysis of easy questions, if a model can recognize increased difficulty, we would expect it to generate longer responses for questions in the Shared-Hard set compared to those in its own advantage set. However, as shown in Table~\ref{tab: t test for hard questions}, we do not observe a clear pattern. Instead, the average token lengths in the Shared-Hard subsets are sometimes shorter than those in each model’s specific advantage set. This suggests that, when the questions are too challenging, models may struggle to be aware of, and respond to, increased difficulty in hard questions. The models may underestimate the difficulty of the problem and respond with unwarranted confidence, or lack the necessary knowledge and capacity to attempt a response whose length reflects the increased complexity of the question.

\begin{figure}[h]
    \centering
    \includegraphics[width=1\linewidth]{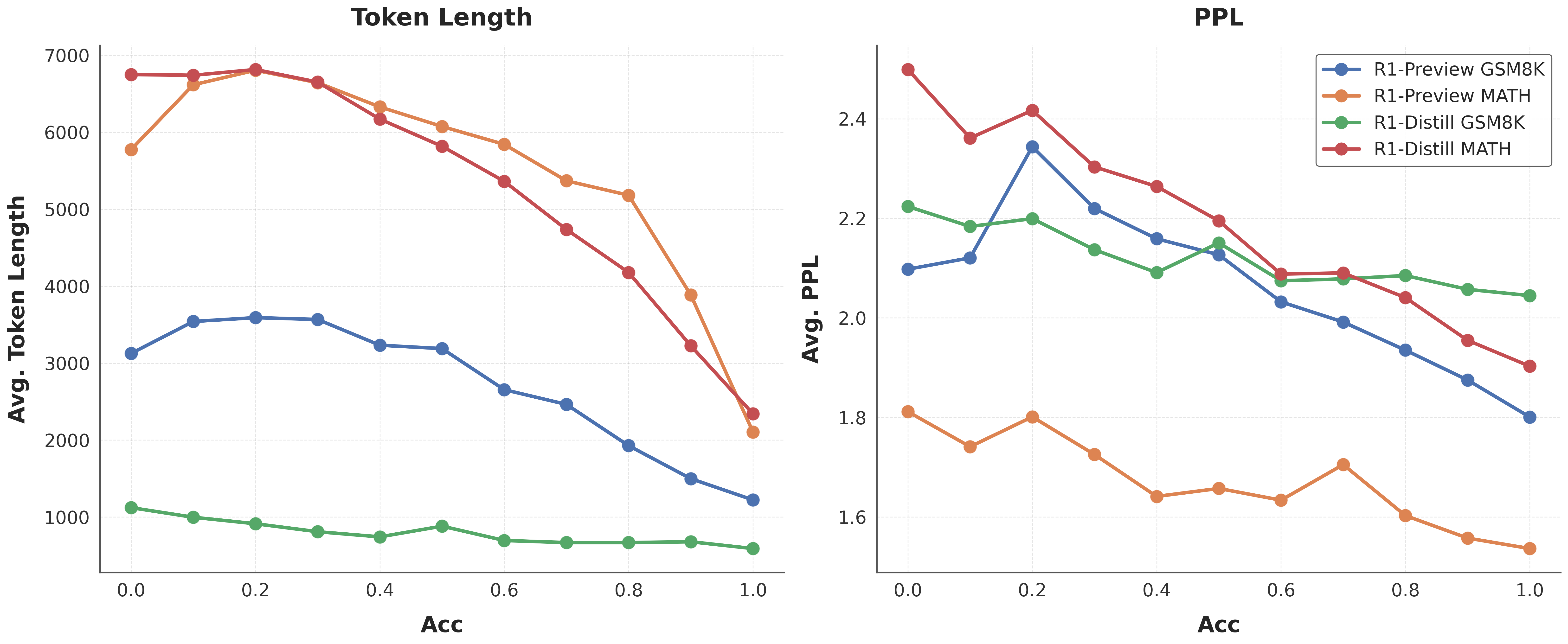}
    \caption{Average token length and perplexity of model responses across questions with different accuracy. }
    \label{fig: acc with token length and perplexity}
\end{figure}

\paragraph{Length and Perplexity Pattern on Question Difficulty} In Figure \ref{fig: acc with token length and perplexity}, we analyze how response length and perplexity vary on questions with different accuracy, which serves as a finer-grained proxy for question difficulty. We observe that average response length initially increases and then decreases as accuracy improves (i.e., as questions become easier), rather than decreasing monotonically—except for R1-Distill on GSM8K. When examining perplexity, we find that for R1-Preview on GSM8K, perplexity first increases as accuracy rises from 0 to 0.2, then decreases, indicating potential overconfidence on these most difficult questions. In contrast, for all other model-dataset pairs, perplexity decreases steadily as accuracy increases, suggesting that models become more confident as questions become easier. These trends in Figure~ \ref{fig: acc with token length and perplexity}  highlight two key failure modes when questions exceed a model’s capability. In Appendix \ref{app: question level additional analysis}, we show the pairwise heatmaps of accuracy, token length and perplexity. We observe that for questions with 0 accuracy, response lengths are more broadly distributed across a wide range of token lengths, rather than concentrated in the high token-length region. In contrast, for questions with higher accuracy, response lengths are more tightly clustered around shorter lengths.

\section{Effect of Length Preference Optimization}
Prior work has explored length preference optimization as a strategy to balance correctness and brevity. A common approach involves generating multiple samples per input and constructing preference pairs that favor correct and/or shorter responses. For example, \citet{yang2025towards} select the shortest correct response as the preferred one and use both the longest incorrect and shortest incorrect responses as negative samples. Similarly, \citet{shen2025dast} propose a dual-pairing scheme: (1) preferring shorter responses among correct answers, and (2) penalizing shorter responses among incorrect ones. Although intuitive, these methods face two key practical limitations: (a) they require generating many samples per input (typically $N \geq 8$), which is often more computationally expensive than the actual fine-tuning process, and the utility of generated samples are low--many samples are filtered out when constructing the preference pair; and (b) constructing these preference pairs requires data with ground-truth answers, limiting their applicability in settings where only unlabeled data is available.  

\begin{figure}[h]
    \centering
    \includegraphics[width=1\linewidth]{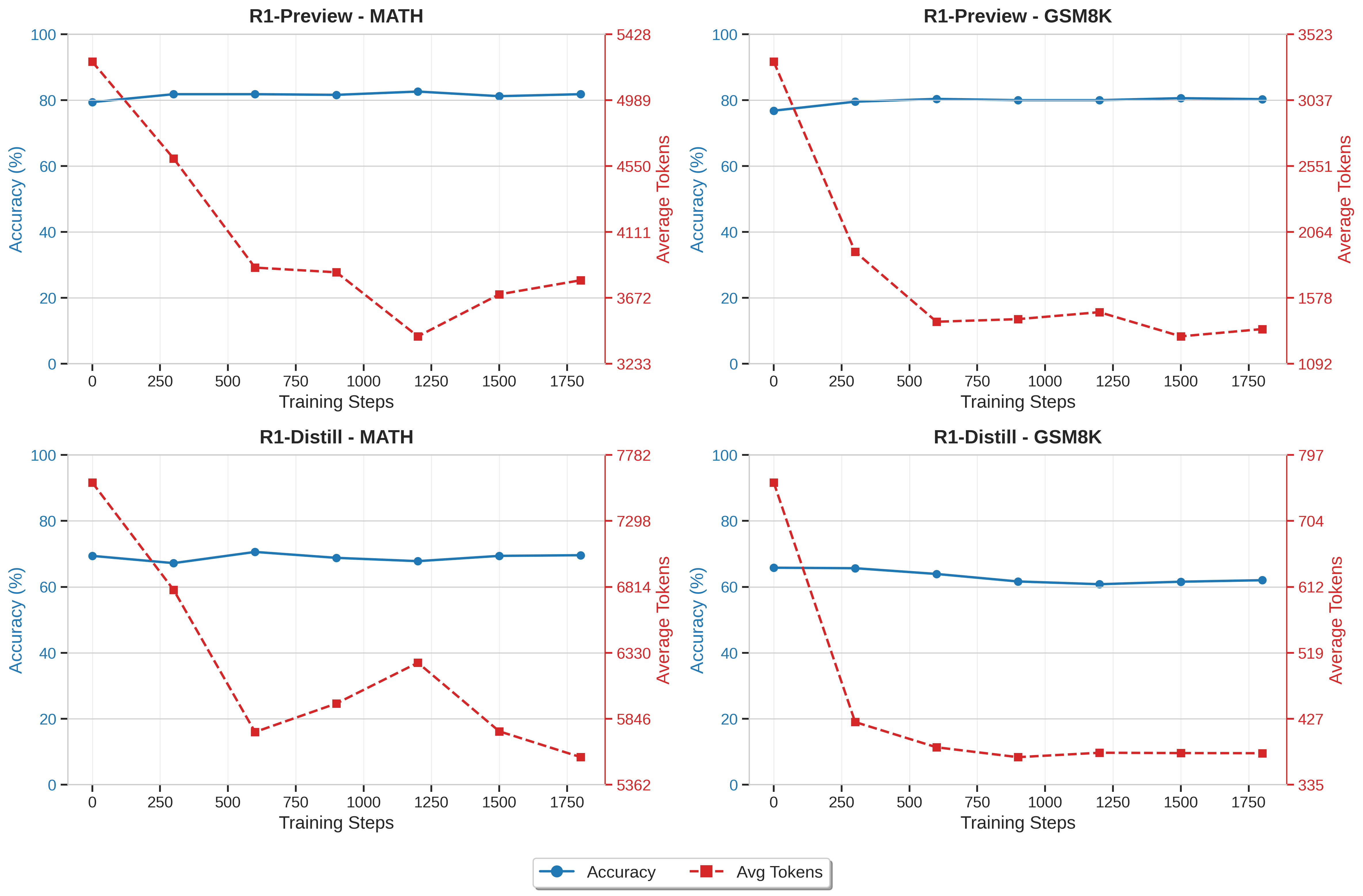}
    \caption{Testing accuracy and average token length when applying the preference optimization algorithm SimPO \cite{meng2024simpo}, trained with preference pairs favoring shorter generations irrespective of correctness. }
    \label{fig: simPO-acc-token}
\end{figure}

Motivated by our earlier analysis—which suggests that shorter generations can already achieve strong performance—we investigate whether it is possible to optimize for shorter outputs without relying on correctness signals during training.
Specifically, we ask: Can preference optimization that solely prioritizes shorter outputs achieve comparable model performance while reducing the generation length, even without access to ground-truth labels? To explore this, we adopt Simple Preference Optimization (SimPO \citealp{meng2024simpo}), an effective offline preference optimization algorithm that uses the average log-probability of a sequence as an implicit reward, with a simplified preference pair construction protocol—we sample just two responses per question and always prefer the shorter one, regardless of correctness. This enables training on entirely unlabeled data while significantly reducing the computational overhead typically associated with large-scale sample generation. We train on the GSM8K and MATH datasets and evaluate performance on their respective test sets. For training data generation, we use a maximum generation length of 8k tokens  to accelerate sampling and set both temperature $T$ and top-$p$ to 1. During testing, we allow up to 32k generation length and apply greedy decoding.

\begin{figure}[h]
    \centering
    \includegraphics[width=1\linewidth]{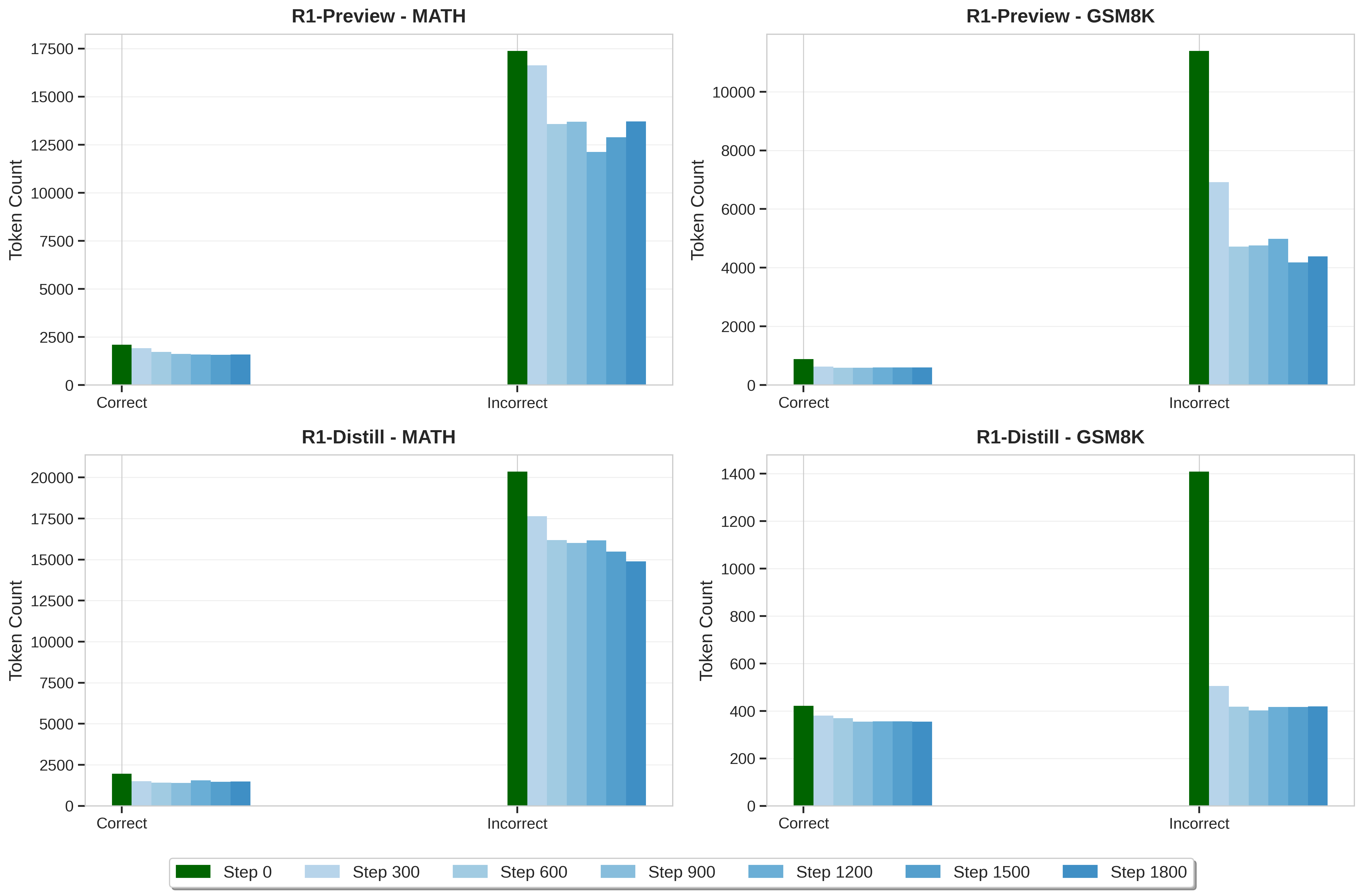}
    \caption{Average token length for correct and incorrect responses across different training steps.}
    \label{fig: simPO-token length bars}
\end{figure}

\paragraph{Preferring shorter responses reduces generation length while maintaining acceptable accuracy. } In Figure \ref{fig: simPO-acc-token}, we show how accuracy and average token length vary over SimPO training steps. We observe that although accuracy fluctuates during training, it remains within an acceptable range without significant degradation, while the average token length decreases substantially—by approximately 30\% to 60\%. While the overall generation length is clearly reduced, it remains unclear whether this reduction is primarily due to shortening correct responses, incorrect ones, or both. To investigate this, we separately track the token length for correct and incorrect responses across training steps, as shown in Figure~\ref{fig: simPO-token length bars}.

\paragraph{Length Reduction  for Incorrect Responses Contributes More to Length Reduction.} Figure \ref{fig: simPO-token length bars} reveals that incorrect responses are generally much longer than correct ones.  Consequently, the observed reduction in overall generation length is mainly driven by shortening the incorrect responses. On the GSM8K dataset, the average token length of incorrect responses drops significantly during the training, while comparably, the length reduction is less prominent for incorrect responses in the MATH dataset.
Additional results, including detailed reduction rates for correct and incorrect responses and trade-offs between accuracy and token length at different training steps, can be found in Appendix~\ref{app: additional results with simpo}.

\section{Conclusion}
In this paper, we empirically study the relationship of generation length and final answer correctness from both sample level and question level.  We find that current reasoning-oriented LLMs struggle to assess problem difficulty and fail to adjust their generation length appropriately for questions they ultimately answer incorrectly—suggesting a tendency toward underthinking. Conversely, for easier problems that they consistently solve correctly, the models are better at estimating difficulty and adjusting response length. However, they still tend to generate unnecessarily long responses in these cases, indicating a pattern of overthinking. As a result, simply preferring shorter generations—without access to correctness signals—may reduce token length without significantly harming performance.
\section{Limitations and Future Work}
To ensure scalability and reduce computational overhead, we use $N=10$ samples per question when analyzing sample-level trends. While this setting aligns with common practice and suffices to reveal consistent patterns, future work could benefit from using larger $N$ to further strengthen statistical reliability and enable finer-grained insights. While our analysis provides new insights into the relationship between generation length and reasoning accuracy, we focus on two widely-used reasoning datasets and two representative reasoning-capable LLM. Extending this analysis to a broader set of models—with different training strategies—and dataset with varying difficulty levels would help assess the generality of our observations and clarify whether these patterns emerge primarily from pretraining or post-training processes such as supervised fine-tuning or reinforcement learning. Thus, we consider it as a future work. 

Our results also motivate practical refinements to test-time inference strategies that aggregate multiple reasoning chains, such as self-consistency \cite{wang2022self}. Incorporating simple heuristics—e.g., deprioritizing excessively long generations—may improve both accuracy and efficiency.

Lastly, the distinct behaviors of models towards easy questions and more challenging questions found in our paper hint at a deeper connection between model calibration and self-correction. Specifically, the effectiveness of self-critique and self-correction may depend on the model's ability to recognize when a problem is within its capabilities. Future work could investigate when self-correction mechanisms succeed, and how they relate to the model’s internal assessment of problem difficulty.

\bibliography{neurips}

\begin{thebibliography}{37}
\providecommand{\natexlab}[1]{#1}
\providecommand{\url}[1]{\texttt{#1}}
\expandafter\ifx\csname urlstyle\endcsname\relax
  \providecommand{\doi}[1]{doi: #1}\else
  \providecommand{\doi}{doi: \begingroup \urlstyle{rm}\Url}\fi

\bibitem[Aggarwal \& Welleck(2025)Aggarwal and Welleck]{aggarwal2025l1}
Pranjal Aggarwal and Sean Welleck.
\newblock L1: Controlling how long a reasoning model thinks with reinforcement learning.
\newblock \emph{arXiv preprint arXiv:2503.04697}, 2025.

\bibitem[Ballon et~al.(2025)Ballon, Algaba, and Ginis]{ballon2025relationship}
Marthe Ballon, Andres Algaba, and Vincent Ginis.
\newblock The relationship between reasoning and performance in large language models--o3 (mini) thinks harder, not longer.
\newblock \emph{arXiv preprint arXiv:2502.15631}, 2025.

\bibitem[Chen et~al.(2025)Chen, Qin, Liu, Peng, Guan, Wang, Hu, Zhou, Gao, and Che]{chen2025towards}
Qiguang Chen, Libo Qin, Jinhao Liu, Dengyun Peng, Jiannan Guan, Peng Wang, Mengkang Hu, Yuhang Zhou, Te~Gao, and Wangxiang Che.
\newblock Towards reasoning era: A survey of long chain-of-thought for reasoning large language models.
\newblock \emph{arXiv preprint arXiv:2503.09567}, 2025.

\bibitem[Chen et~al.(2024)Chen, Xu, Liang, He, Pang, Yu, Song, Liu, Zhou, Zhang, et~al.]{chen2024not}
Xingyu Chen, Jiahao Xu, Tian Liang, Zhiwei He, Jianhui Pang, Dian Yu, Linfeng Song, Qiuzhi Liu, Mengfei Zhou, Zhuosheng Zhang, et~al.
\newblock Do not think that much for 2+ 3=? on the overthinking of o1-like llms.
\newblock \emph{arXiv preprint arXiv:2412.21187}, 2024.

\bibitem[Cobbe et~al.(2021)Cobbe, Kosaraju, Bavarian, Chen, Jun, Kaiser, Plappert, Tworek, Hilton, Nakano, Hesse, and Schulman]{cobbe2021gsm8k}
Karl Cobbe, Vineet Kosaraju, Mohammad Bavarian, Mark Chen, Heewoo Jun, Lukasz Kaiser, Matthias Plappert, Jerry Tworek, Jacob Hilton, Reiichiro Nakano, Christopher Hesse, and John Schulman.
\newblock Training verifiers to solve math word problems.
\newblock \emph{arXiv preprint arXiv:2110.14168}, 2021.

\bibitem[Damani et~al.(2024)Damani, Shenfeld, Peng, Bobu, and Andreas]{damani2024learning}
Mehul Damani, Idan Shenfeld, Andi Peng, Andreea Bobu, and Jacob Andreas.
\newblock Learning how hard to think: Input-adaptive allocation of lm computation.
\newblock \emph{arXiv preprint arXiv:2410.04707}, 2024.

\bibitem[DeepSeek-AI(2025)]{deepseekai2025deepseekr1incentivizingreasoningcapability}
DeepSeek-AI.
\newblock Deepseek-r1: Incentivizing reasoning capability in llms via reinforcement learning, 2025.
\newblock URL \url{https://arxiv.org/abs/2501.12948}.

\bibitem[Fu et~al.()Fu, Chen, Zhuang, Fu, Stoica, and Zhang]{fu2025reasoning}
Yichao Fu, Junda Chen, Yonghao Zhuang, Zheyu Fu, Ion Stoica, and Hao Zhang.
\newblock Reasoning without self-doubt: More efficient chain-of-thought through certainty probing.
\newblock In \emph{ICLR 2025 Workshop on Foundation Models in the Wild}.

\bibitem[Google(2024)]{Flash-thinking}
Google.
\newblock Gemini 2.0 flash thinking mode.
\newblock \emph{https://ai.google.dev/ gemini-api/docs/thinking-mode}, 2024.

\bibitem[Guo et~al.(2025)Guo, Yang, Zhang, Song, Zhang, Xu, Zhu, Ma, Wang, Bi, et~al.]{guo2025deepseek}
Daya Guo, Dejian Yang, Haowei Zhang, Junxiao Song, Ruoyu Zhang, Runxin Xu, Qihao Zhu, Shirong Ma, Peiyi Wang, Xiao Bi, et~al.
\newblock Deepseek-r1: Incentivizing reasoning capability in llms via reinforcement learning.
\newblock \emph{arXiv preprint arXiv:2501.12948}, 2025.

\bibitem[Han et~al.(2024)Han, Wang, Fang, Zhao, Ma, and Chen]{han2024token}
Tingxu Han, Zhenting Wang, Chunrong Fang, Shiyu Zhao, Shiqing Ma, and Zhenyu Chen.
\newblock Token-budget-aware llm reasoning.
\newblock \emph{arXiv preprint arXiv:2412.18547}, 2024.

\bibitem[Hendrycks et~al.(2021)Hendrycks, Burns, Kadavath, Arora, Basart, Tang, Song, and Steinhardt]{hendrycks2021measuring}
Dan Hendrycks, Collin Burns, Saurav Kadavath, Akul Arora, Steven Basart, Eric Tang, Dawn Song, and Jacob Steinhardt.
\newblock Measuring mathematical problem solving with the math dataset.
\newblock \emph{arXiv preprint arXiv:2103.03874}, 2021.

\bibitem[Jin et~al.(2024)Jin, Yu, Shu, Zhao, Hua, Meng, Zhang, and Du]{jin2024impact}
Mingyu Jin, Qinkai Yu, Dong Shu, Haiyan Zhao, Wenyue Hua, Yanda Meng, Yongfeng Zhang, and Mengnan Du.
\newblock The impact of reasoning step length on large language models.
\newblock \emph{arXiv preprint arXiv:2401.04925}, 2024.

\bibitem[Li et~al.(2024)Li, Yuan, Feng, Pan, Wang, Sun, Wang, and Li]{li2024escape}
Yiwei Li, Peiwen Yuan, Shaoxiong Feng, Boyuan Pan, Xinglin Wang, Bin Sun, Heda Wang, and Kan Li.
\newblock Escape sky-high cost: Early-stopping self-consistency for multi-step reasoning.
\newblock \emph{arXiv preprint arXiv:2401.10480}, 2024.

\bibitem[Li et~al.(2025)Li, Zhang, Zhang, Zhang, Liu, Yao, Xu, Zheng, Wang, Chen, et~al.]{li2025system}
Zhong-Zhi Li, Duzhen Zhang, Ming-Liang Zhang, Jiaxin Zhang, Zengyan Liu, Yuxuan Yao, Haotian Xu, Junhao Zheng, Pei-Jie Wang, Xiuyi Chen, et~al.
\newblock From system 1 to system 2: A survey of reasoning large language models.
\newblock \emph{arXiv preprint arXiv:2502.17419}, 2025.

\bibitem[Luo et~al.(2025{\natexlab{a}})Luo, Shen, He, Wang, Liu, Li, Tan, Cao, and Tao]{luo2025o1}
Haotian Luo, Li~Shen, Haiying He, Yibo Wang, Shiwei Liu, Wei Li, Naiqiang Tan, Xiaochun Cao, and Dacheng Tao.
\newblock O1-pruner: Length-harmonizing fine-tuning for o1-like reasoning pruning.
\newblock \emph{arXiv preprint arXiv:2501.12570}, 2025{\natexlab{a}}.

\bibitem[Luo et~al.(2025{\natexlab{b}})Luo, Tan, Wong, Shi, Tang, Roongta, Cai, Luo, Zhang, Li, Popa, and Stoica]{deepscaler2025}
Michael Luo, Sijun Tan, Justin Wong, Xiaoxiang Shi, William~Y. Tang, Manan Roongta, Colin Cai, Jeffrey Luo, Tianjun Zhang, Li~Erran Li, Raluca~Ada Popa, and Ion Stoica.
\newblock Deepscaler: Surpassing o1-preview with a 1.5b model by scaling rl.
\newblock \url{https://pretty-radio-b75.notion.site/DeepScaleR-Surpassing-O1-Preview-with-a-1-5B-Model-by-Scaling-RL}, 2025{\natexlab{b}}.

\bibitem[Ma et~al.(2025{\natexlab{a}})Ma, He, Snell, Griggs, Min, and Zaharia]{ma2025reasoning}
Wenjie Ma, Jingxuan He, Charlie Snell, Tyler Griggs, Sewon Min, and Matei Zaharia.
\newblock Reasoning models can be effective without thinking.
\newblock \emph{arXiv preprint arXiv:2504.09858}, 2025{\natexlab{a}}.

\bibitem[Ma et~al.(2025{\natexlab{b}})Ma, Wan, Yu, Fang, and Wang]{ma2025cot}
Xinyin Ma, Guangnian Wan, Runpeng Yu, Gongfan Fang, and Xinchao Wang.
\newblock Cot-valve: Length-compressible chain-of-thought tuning.
\newblock \emph{arXiv preprint arXiv:2502.09601}, 2025{\natexlab{b}}.

\bibitem[Manvi et~al.(2024)Manvi, Singh, and Ermon]{manvi2024adaptive}
Rohin Manvi, Anikait Singh, and Stefano Ermon.
\newblock Adaptive inference-time compute: Llms can predict if they can do better, even mid-generation.
\newblock \emph{arXiv preprint arXiv:2410.02725}, 2024.

\bibitem[Meng et~al.(2024)Meng, Xia, and Chen]{meng2024simpo}
Yu~Meng, Mengzhou Xia, and Danqi Chen.
\newblock Simpo: Simple preference optimization with a reference-free reward.
\newblock \emph{Advances in Neural Information Processing Systems}, 37:\penalty0 124198--124235, 2024.

\bibitem[Muennighoff et~al.(2025)Muennighoff, Yang, Shi, Li, Fei-Fei, Hajishirzi, Zettlemoyer, Liang, Cand{\`e}s, and Hashimoto]{muennighoff2025s1}
Niklas Muennighoff, Zitong Yang, Weijia Shi, Xiang~Lisa Li, Li~Fei-Fei, Hannaneh Hajishirzi, Luke Zettlemoyer, Percy Liang, Emmanuel Cand{\`e}s, and Tatsunori Hashimoto.
\newblock s1: Simple test-time scaling.
\newblock \emph{arXiv preprint arXiv:2501.19393}, 2025.

\bibitem[Munkhbat et~al.(2025)Munkhbat, Ho, Kim, Yang, Kim, and Yun]{munkhbat2025self}
Tergel Munkhbat, Namgyu Ho, Seohyun Kim, Yongjin Yang, Yujin Kim, and Se-Young Yun.
\newblock Self-training elicits concise reasoning in large language models.
\newblock \emph{arXiv preprint arXiv:2502.20122}, 2025.

\bibitem[OpenAI(2024)]{openAI-o1}
OpenAI.
\newblock Learning to reason with llms.
\newblock \emph{https://openai.com/index/ learning-to-reason-with-llms}, 2024.

\bibitem[Pan et~al.(2024)Pan, Zhang, Zhang, Liu, Wang, and Li]{pan2024dynathink}
Jiabao Pan, Yan Zhang, Chen Zhang, Zuozhu Liu, Hongwei Wang, and Haizhou Li.
\newblock Dynathink: Fast or slow? a dynamic decision-making framework for large language models.
\newblock \emph{arXiv preprint arXiv:2407.01009}, 2024.

\bibitem[Qu et~al.(2025)Qu, Yang, Setlur, Tunstall, Beeching, Salakhutdinov, and Kumar]{qu2025optimizing}
Yuxiao Qu, Matthew~YR Yang, Amrith Setlur, Lewis Tunstall, Edward~Emanuel Beeching, Ruslan Salakhutdinov, and Aviral Kumar.
\newblock Optimizing test-time compute via meta reinforcement fine-tuning.
\newblock \emph{arXiv preprint arXiv:2503.07572}, 2025.

\bibitem[Qwen(2024)]{QwQ}
Qwen.
\newblock Qwq: Reflect deeply on the boundaries of the unknown.
\newblock \emph{https://qwenlm.github.io/blog/qwq-32b-preview/}, 2024.

\bibitem[Shen et~al.(2025)Shen, Zhang, Huang, Shi, Zhang, Yan, Wang, Wang, and Lian]{shen2025dast}
Yi~Shen, Jian Zhang, Jieyun Huang, Shuming Shi, Wenjing Zhang, Jiangze Yan, Ning Wang, Kai Wang, and Shiguo Lian.
\newblock Dast: Difficulty-adaptive slow-thinking for large reasoning models.
\newblock \emph{arXiv preprint arXiv:2503.04472}, 2025.

\bibitem[Team et~al.(2025)Team, Du, Gao, Xing, Jiang, Chen, Li, Xiao, Du, Liao, et~al.]{team2025kimi}
Kimi Team, Angang Du, Bofei Gao, Bowei Xing, Changjiu Jiang, Cheng Chen, Cheng Li, Chenjun Xiao, Chenzhuang Du, Chonghua Liao, et~al.
\newblock Kimi k1. 5: Scaling reinforcement learning with llms.
\newblock \emph{arXiv preprint arXiv:2501.12599}, 2025.

\bibitem[Wang et~al.(2024)Wang, Feng, Li, Yuan, Zhang, Tan, Pan, Hu, and Li]{wang2024make}
Xinglin Wang, Shaoxiong Feng, Yiwei Li, Peiwen Yuan, Yueqi Zhang, Chuyi Tan, Boyuan Pan, Yao Hu, and Kan Li.
\newblock Make every penny count: Difficulty-adaptive self-consistency for cost-efficient reasoning.
\newblock \emph{arXiv preprint arXiv:2408.13457}, 2024.

\bibitem[Wang et~al.(2022)Wang, Wei, Schuurmans, Le, Chi, Narang, Chowdhery, and Zhou]{wang2022self}
Xuezhi Wang, Jason Wei, Dale Schuurmans, Quoc Le, Ed~Chi, Sharan Narang, Aakanksha Chowdhery, and Denny Zhou.
\newblock Self-consistency improves chain of thought reasoning in language models.
\newblock \emph{arXiv preprint arXiv:2203.11171}, 2022.

\bibitem[Wu et~al.(2025)Wu, Wang, Du, Jegelka, and Wang]{wu2025more}
Yuyang Wu, Yifei Wang, Tianqi Du, Stefanie Jegelka, and Yisen Wang.
\newblock When more is less: Understanding chain-of-thought length in llms.
\newblock \emph{arXiv preprint arXiv:2502.07266}, 2025.

\bibitem[Xie et~al.(2025)Xie, Gao, Ren, Luo, Hong, Dai, Zhou, Qiu, Wu, and Luo]{xie2025logic}
Tian Xie, Zitian Gao, Qingnan Ren, Haoming Luo, Yuqian Hong, Bryan Dai, Joey Zhou, Kai Qiu, Zhirong Wu, and Chong Luo.
\newblock Logic-rl: Unleashing llm reasoning with rule-based reinforcement learning.
\newblock \emph{arXiv preprint arXiv:2502.14768}, 2025.

\bibitem[Xu et~al.(2024)Xu, Li, Sun, and Qian]{xu2024adaption}
Mayi Xu, Yongqi Li, Ke~Sun, and Tieyun Qian.
\newblock Adaption-of-thought: Learning question difficulty improves large language models for reasoning.
\newblock In \emph{Proceedings of the 2024 Conference on Empirical Methods in Natural Language Processing}, pp.\  5468--5495, 2024.

\bibitem[Yang et~al.(2025)Yang, Ma, Lin, and Wei]{yang2025towards}
Wenkai Yang, Shuming Ma, Yankai Lin, and Furu Wei.
\newblock Towards thinking-optimal scaling of test-time compute for llm reasoning.
\newblock \emph{arXiv preprint arXiv:2502.18080}, 2025.

\bibitem[Yeo et~al.(2025)Yeo, Tong, Niu, Neubig, and Yue]{yeo2025demystifying}
Edward Yeo, Yuxuan Tong, Morry Niu, Graham Neubig, and Xiang Yue.
\newblock Demystifying long chain-of-thought reasoning in llms.
\newblock \emph{arXiv preprint arXiv:2502.03373}, 2025.

\bibitem[Zhang \& Zuo(2025)Zhang and Zuo]{zhang2025grpo}
Jixiao Zhang and Chunsheng Zuo.
\newblock Grpo-lead: A difficulty-aware reinforcement learning approach for concise mathematical reasoning in language models.
\newblock \emph{arXiv preprint arXiv:2504.09696}, 2025.

\end{thebibliography}
\bibliographystyle{colm}

\newpage
\appendix

\section{Experimental Details}
\subsection{Prompt}
\begin{figure}[htbp]
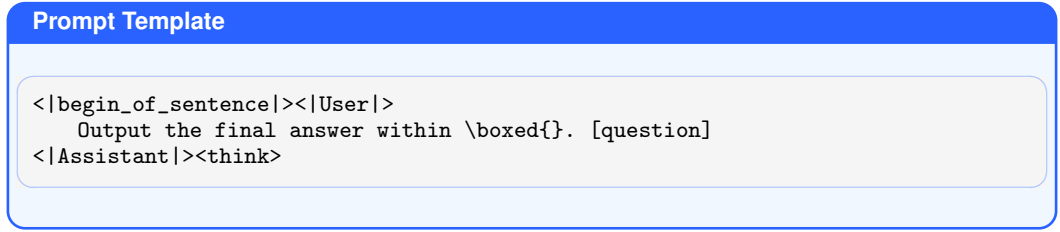

  \centering
  \begin{tcolorbox}[
    colback=lightblue,
    colframe=promptblue,
    boxrule=1pt,
    arc=6pt,
    left=6pt, right=6pt, top=6pt, bottom=6pt,
    title={\textbf{Prompt Template}},
    fonttitle=\sffamily\bfseries\small,
    coltitle=white,
    colbacktitle=promptblue
  ]
    \begin{lstlisting}[
      basicstyle=\ttfamily\small,
      backgroundcolor=\color{codegray},
      columns=fullflexible,
      breaklines=true,
      frame=single,
      framesep=5pt,
      frameround=tttt,
      rulecolor=\color{promptblue!40}
    ]
<|begin_of_sentence|><|User|>
    Output the final answer within \boxed{}. [question]
<|Assistant|><think>
    \end{lstlisting}
  \end{tcolorbox}
  \caption{The prompt template used to generate responses}
  \label{fig:prompt-template}
\end{figure}

The prompt template used to generate the data is given in Figure \ref{fig:prompt-template}.
\subsection{Metrics}
\paragraph{Sample-Level Metrics}
To investigate how correctness varies with length in sample level, we sort the $N$ samples for each question in ascending order of reasoning length $l_i^{(q)}$, creating a length-based ranking. For each rank position $r\in \{1, \cdots, N-1\}$, define the mean reasoning length and mean accuracy at rank $r$ across the dataset:
\begin{equation*}
L_{r}=\frac{1}{|D|}\sum_{q\in D}l_r^{(q)}, ~\text{Acc}_r=\frac{1}{|D|}\sum_{q\in D}{c_r^{(q)}}
\end{equation*}
These metrics allow us to examine whether, across different reasoning paths for the same question, shorter or longer responses tend to yield higher accuracy.

\paragraph{Question-Level Metrics}
At the question level, we study the relationship between a question’s overall average reasoning length and its overall correctness averaged over $N$ samples. Specifically, for each question $q$, we compute:
\begin{equation*}
L^{(q)} = \frac{1}{N}\sum_{i\in [N]}l_i^{(q)}, \text{Acc}^{(q)}=\frac{1}{N}\sum_{i\in [N]}c_i^{(q)}
\end{equation*}

When accessing how overall response length relates to correctness (as Table \ref{tab:token_length_correlations}), we use the average token lengths of correct and incorrect responses, denoted as $L^{\checkmark}$ and and $L^{\times}$, respectively:
\begin{equation*}
L^{\checkmark} =\frac{1}{|D^{\checkmark}|} \sum_{q\in D^{\checkmark}}l^{(q)}, ~L^{\times} =\frac{1}{|D^{\times}|} \sum_{q\in D^{\times}}l^{(q)},
\end{equation*}
where $l^{(q)}$ is the response length for question $q$, and $D^{\checkmark}$ and $D^{\times}$ are the sets of questions answered correctly and incorrectly, respectively.

\section{Additional Experiments for Sample Level Analysis}\label{app: sample level additional analysis}
\subsection{Distribution of Extremes of Rank Indices.}

\begin{figure}[h]
    \centering
    \includegraphics[width=1\linewidth]{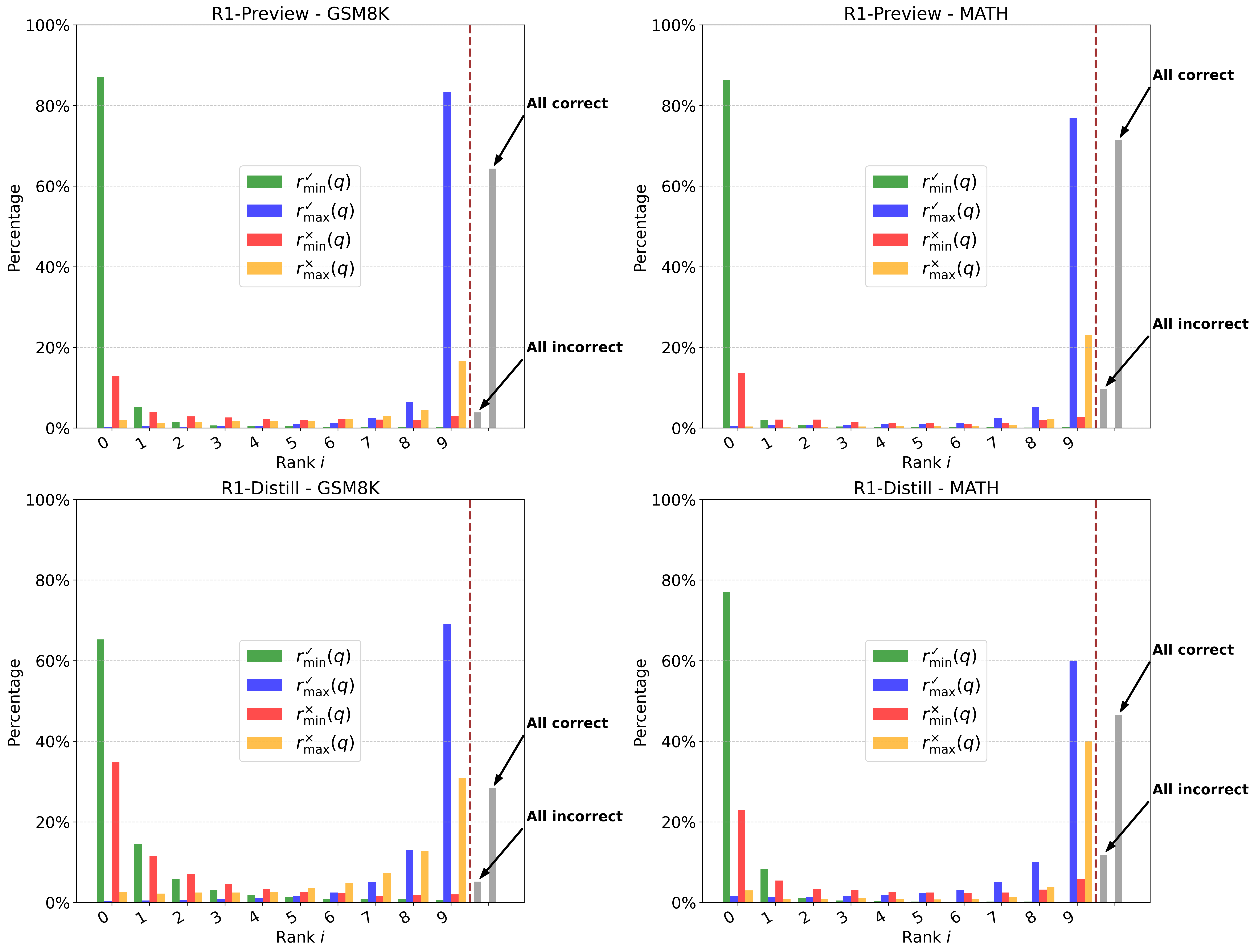}
    \caption{Distribution of the extreme rank indices $r_{\min}^{\checkmark}, r_{\max}^{\checkmark}, r_{\min}^{\times}, r_{\max}^{\times}$. These represent the positions of the shortest and longest correct and incorrect samples in the length-ranked list for each question. The gray bar on the right indicates the proportion of questions for which all samples are either correct or incorrect—cases where the corresponding extreme rank indices do not exist.}
    \label{fig: first correct index}
\end{figure}

To analyze the distribution of correctness at the extremes of the ranked list for each question, we define the following rank-based indicators:

\begin{itemize}
\item$r_{\min}^{\checkmark}(q)$: rank of the shortest correct sample, i.e., $\min\{r|c_r^{(q)}=1\}$.
\item  $r_{\max}^{\checkmark}(q)$: rank of the longest correct sample, i.e., $\max\{r|c_r^{(q)}=1\}$.
\item $r_{\min}^{\times}(q)$: rank of the shortest incorrect sample, i.e., $\min\{r|c_r^{(q)}=0\}$.
\item  $r_{\max}^{\times}(q)$: rank of the longest incorrect sample, , i.e., $\max\{r|c_r^{(q)}=0\}$.
\end{itemize}

$r_{\min}^{\checkmark}(q) = i$ indicates that the $i$-th shortest sample already contains sufficient reasoning to solve the problem correctly. In contrast, $r_{\max}^{\checkmark}(q) = i$ identifies the longest reasoning length for which correctness is still preserved. When $r_{\max}^{\checkmark}(q) < N-1$, it suggests that any further increase in reasoning length—beyond the $i$-th length ranked sample—introduces noise or errors that begin to degrade performance. 
In this sense, $r_{\max}^{\checkmark}(q)$ helps locate the boundary at which reasoning becomes too excessive and starts to erode correctness. Similarly, the distributions of $r_{\min}^{\times}(q) = i$ and $r_{\max}^{\times}(q) = i$ characterize the boundaries of incorrect responses.
These distribution of these extremes of rank indices are given in Figure \ref{fig: first correct index}. 

\begin{figure}[h]
    \centering
    \includegraphics[width=1\linewidth]{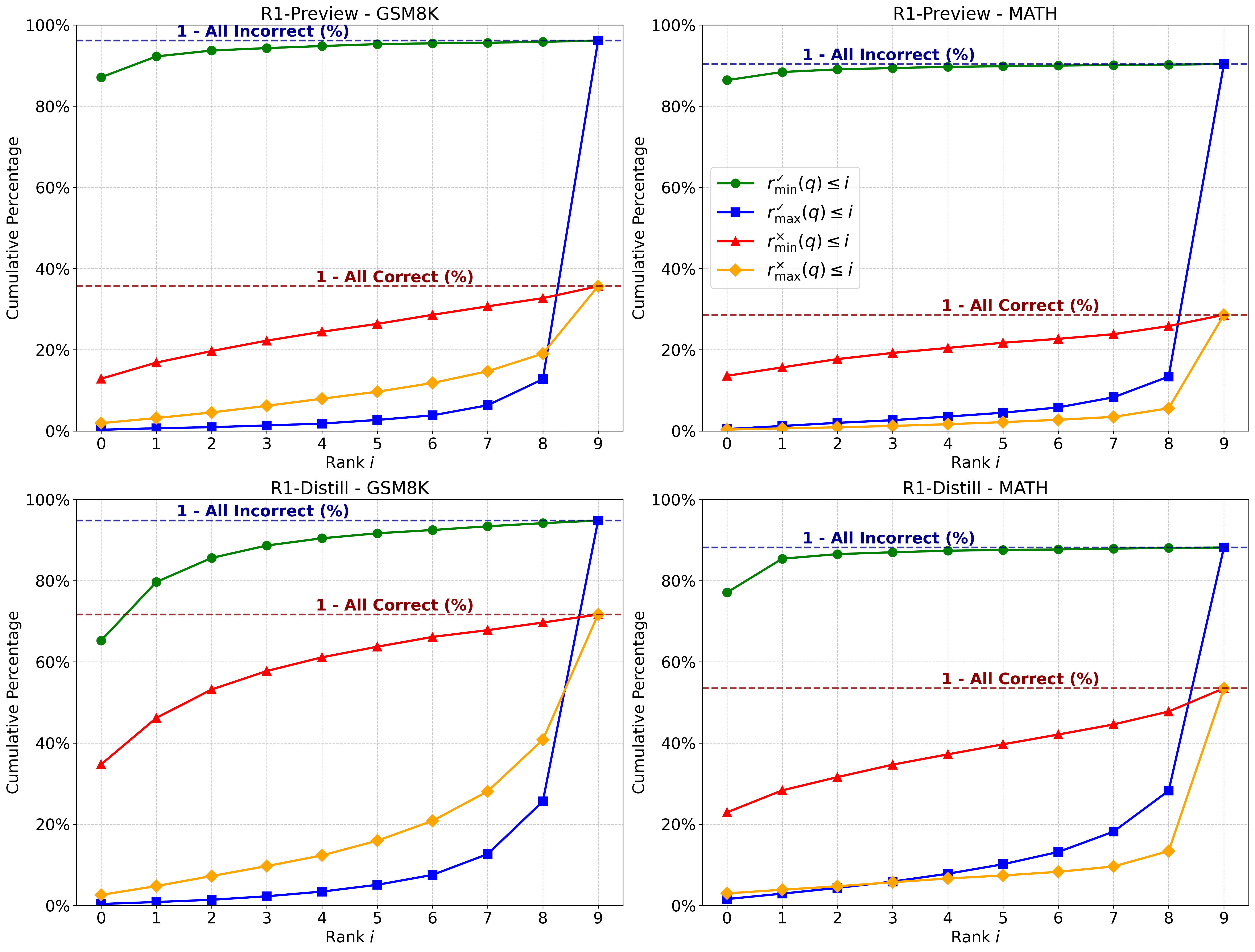}
    \caption{Cumulative distribution the extreme rank indices $r_{\min}^{\checkmark}, r_{\max}^{\checkmark}, r_{\min}^{\times}, r_{\max}^{\times}$. The dashed line labeled "1 - All Incorrect" indicates the proportion of questions with at least one correct sample, while the line labeled "1 - All Correct" marks the proportion of questions where at least one sample is incorrect.}
    \label{fig: cumulative distribution of extreme}
\end{figure}

\paragraph{Correct Answers Often Suffice Within the Shortest Few Samples}
As shown in Figure~\ref{fig: first correct index}, over 80\% of questions on both GSM8K and MATH when using R1-Preview, the shortest sample is already correct, i.e., $r_{\min}^{\checkmark}(q) = 0$. For R1-Distill, this proportion is slightly lower—over 60\% for GSM8K and over 70\% for MATH. Focusing on the distribution of $r_{\min}^{\checkmark}(q)$ among questions where $r_{\min}^{\checkmark}(q)$ exist, (i.e., at least one sample is correct), we observe that for the R1-Preview-GSM8K setting, most questions are answered correctly by either the shortest or second shortest sample, with $r_{\min}^{\checkmark}(q) = 0$ and $1$ together covering a large majority. For R1-Preview-MATH, $r_{\min}^{\checkmark}(q) = 0$ alone accounts for the majority of answerable questions. In contrast, for R1-Distill-GSM8K, $r_{\min}^{\checkmark}(q) = 0$ represents a smaller proportion compared to the other three settings, and a non-negligible fraction of questions require slightly longer reasoning, with $r_{\min}^{\checkmark}(q) = 1$ and $2$ also contributing substantially. The cumulative distribution of $r^{\checkmark}_{\min} \leq i$ in Figure \ref{fig: cumulative distribution of extreme} shows the above observation more clearly. Specifically, the plot for $r^{\checkmark}_{\min} \leq i$ in Figure \ref{fig: cumulative distribution of extreme} shows the proportion of questions that are answered correctly by the $i$-th shortest or shorter response. For all settings except R1-Distill–GSM8K, the cumulative percentage increases rapidly and begins to plateau after $i > 1$, indicating that the top two shortest generations are sufficient to answer nearly all questions the model is capable of solving. 

\paragraph{Excessive Reasoning Can Degrade Performance} The distribution of $r_{\max}^{\checkmark}(q)$ highlights the point at which increasing reasoning length begins to undermine correctness.  Specifically, when $r_{\max}^{\checkmark}(q) = i$ for any $i < N-1$, it indicates that correctness is maintained up to the $i$-th ranked sample but is lost at rank $i{+}1$. This suggests that beyond a certain length, additional reasoning introduces errors that compromise an otherwise correct answer. This effect is most apparent near the end of the ranked list—that is, among the longest few samples. For R1-Preview, the proportion of questions with $r_{\max}^{\checkmark}(q) = i$ remains low for $i < N{-}2$, but becomes noticeable at $i = N{-}2$: approximately 5\% of questions that are answered correctly at this rank become incorrect at the longer response. For R1-Distill, the issue is more pronounced—around 10\% of questions that are correct at rank $N-2$ become incorrect at rank $N-1$. These patterns suggest that excessive reasoning can degrade performance by either exceeding the maximum token length or introducing errors that turn an initially correct answer into an incorrect one.

\subsection{Correlation of Token Length of Each Rank and the Correctness}
\begin{figure}[h]
    \centering
    \includegraphics[width=1\linewidth]{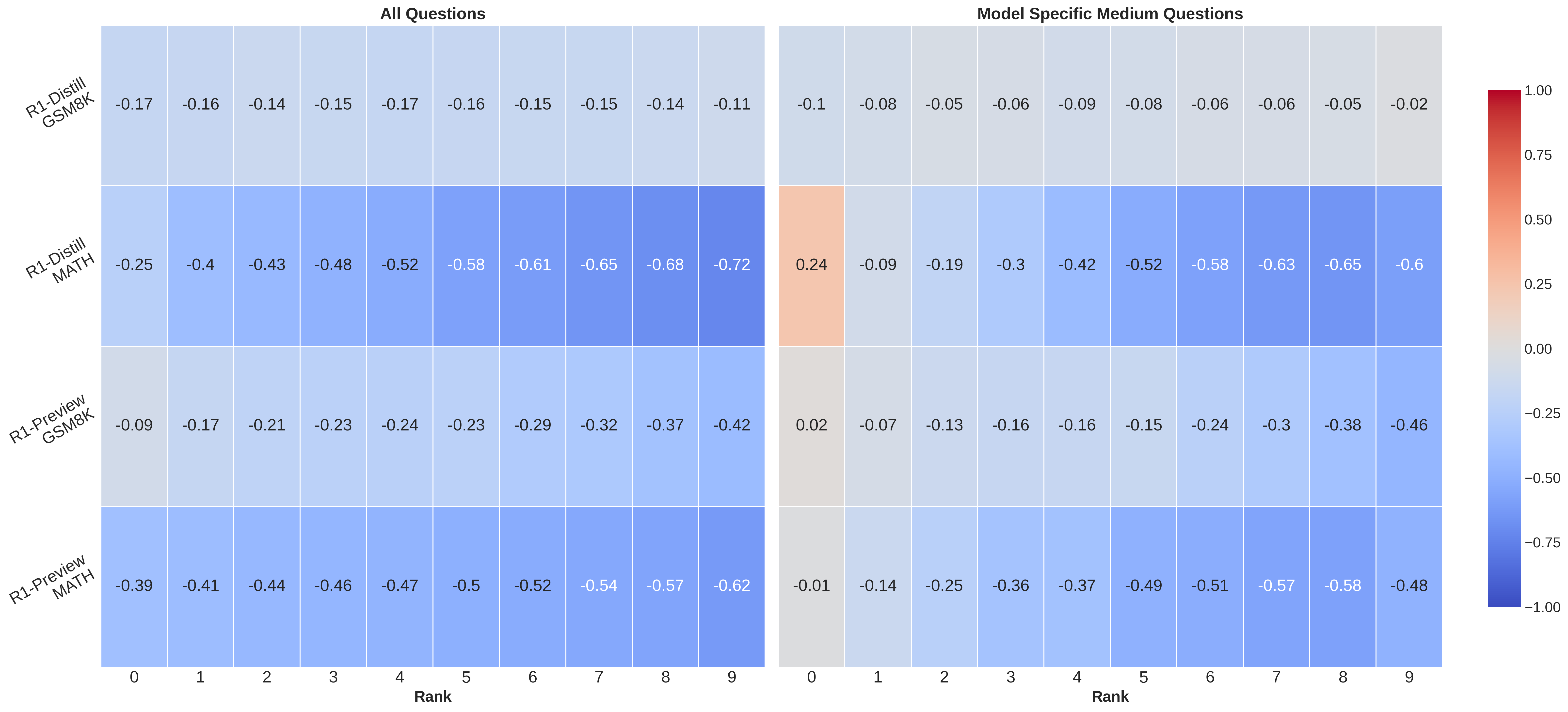}
    \caption{Spearman correlation of answering length and correctness.}
    \label{fig: spearman correlation}
\end{figure}

\begin{figure}[h]
    \centering
    \includegraphics[width=1\linewidth]{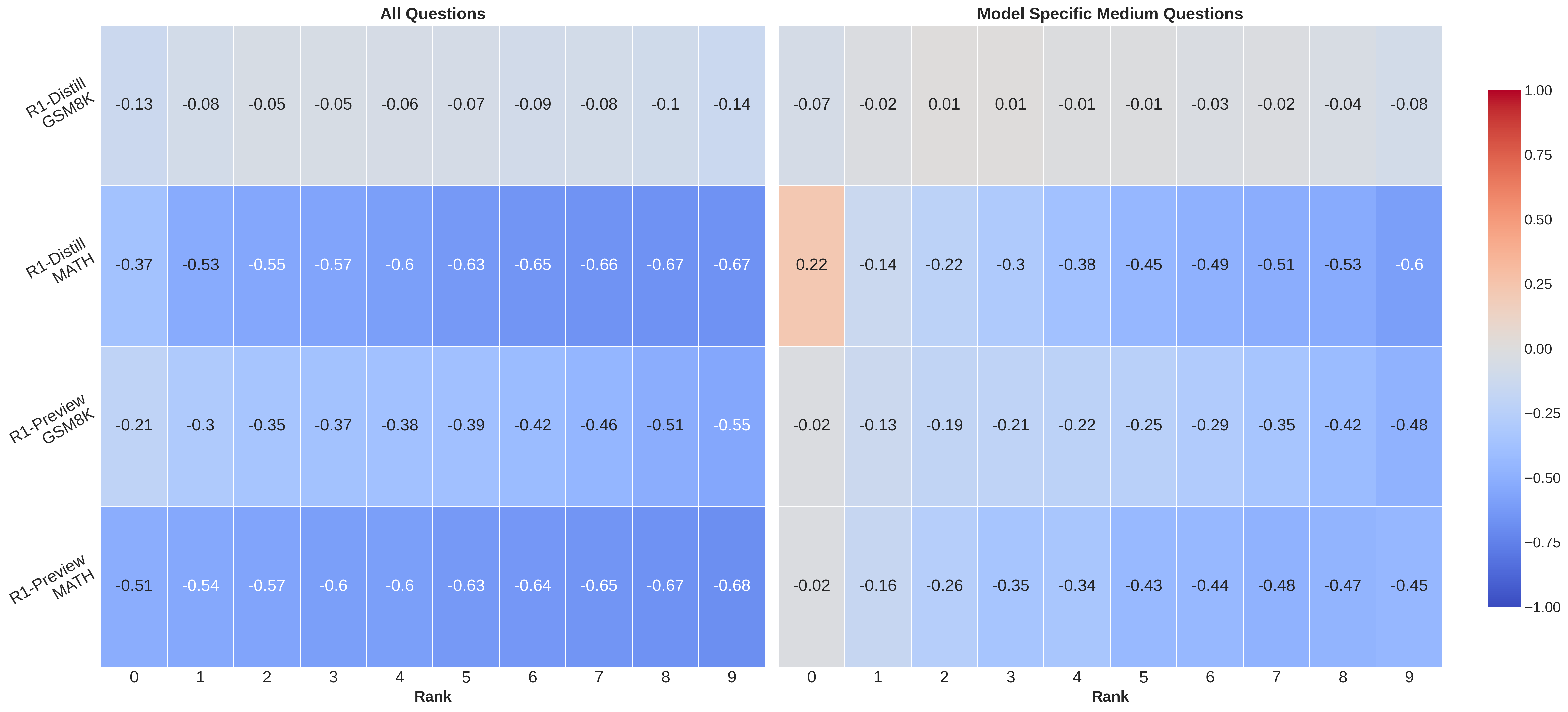}
    \caption{Pearson correlation of answering length and correctness.}
    \label{fig: pearson correlation}
\end{figure}

In Figure \ref{fig: spearman correlation}, we show the spearman correlation of token length and correctness for responses at each rank, the correlations evaluated on all the questions are on the left subgraph while on the right of the subfigure, we evaluated all the medium questions, i.e., questions that are neither all correct nor all incorrect across all the $N$ samples. We find that, there is a strong correlation with correctness and the length among different samples for the same questions. This indicates that, fixing a question difficulty and the model, where the optimal token required to answer the question would be decided, generating more tokens over the required optimal token length would not be helpful. We also computed Pearson correlation in Figure \ref{fig: pearson correlation}, which shows the same trend. 
\section{Statistics for Each Model Specific Difficulty Level}\label{app: statistics for each difficulty level}
\begin{table}[h]
\centering
\caption{The number of questions in each model-specific difficulty level. }
\label{tab:question-distribution}
\begin{tabular}{llrrrr}
\hline
\textbf{Dataset} & \textbf{Model} & \textbf{Easy} & \textbf{Medium} & \textbf{Hard} & \textbf{Total} \\
\hline
\multirow{2}{*}{GSM8K} 
& R1-Preview        & 4809 & 2377 & 287 & \multirow{2}{*}{\textbf{7473}} \\
& R1-Distill& 2118 & 4967 & 388 & \\
\hline
\multirow{2}{*}{MATH} 
& R1-Preview        & 5354 & 1424 & 722 & \multirow{2}{*}{\textbf{7500}} \\
& R1-Distill & 3489 & 3122 & 889 & \\
\hline
\end{tabular}
\end{table}
\begin{figure}[h]
    \centering
    \includegraphics[width=1\linewidth]{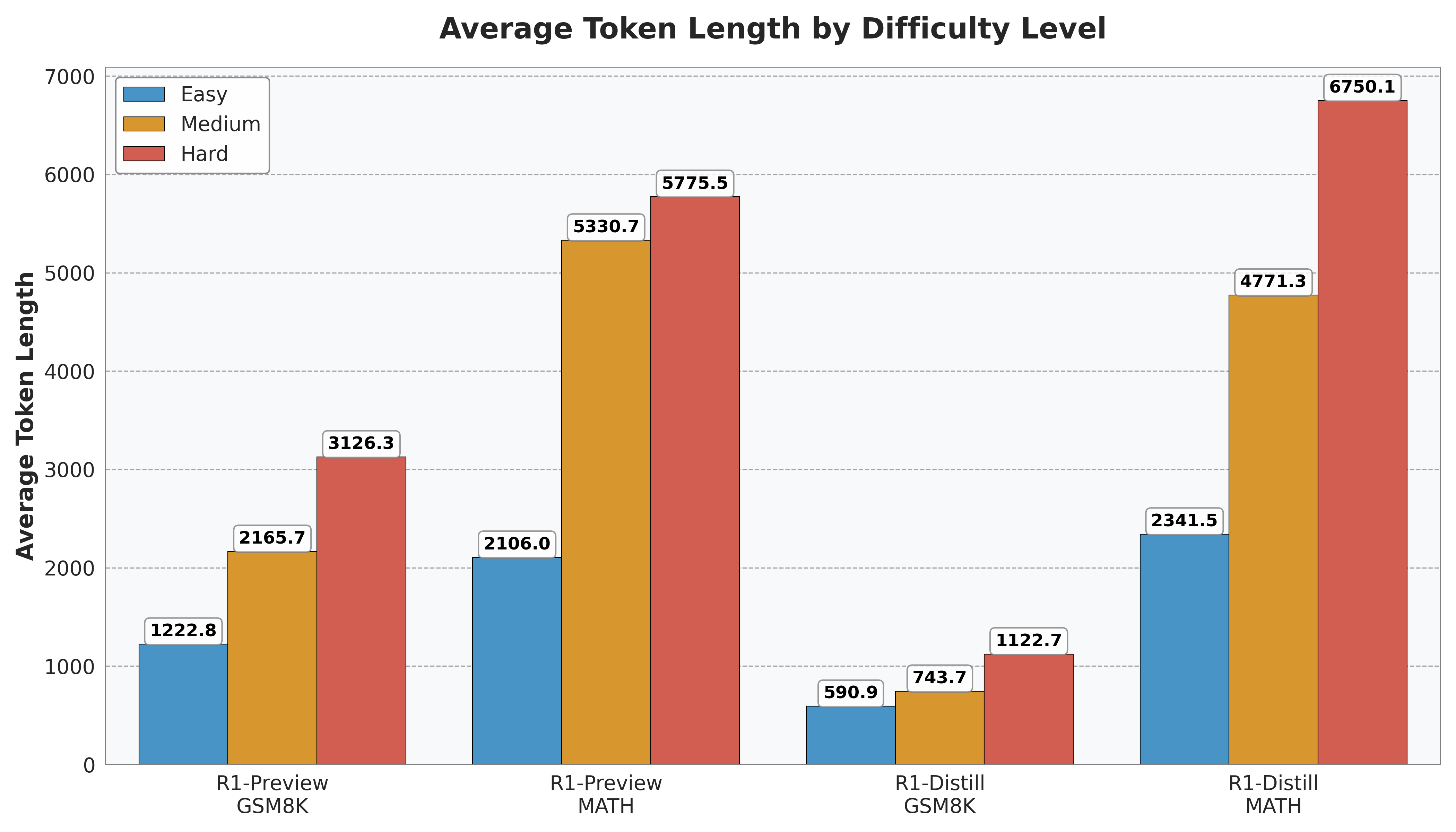}
    \caption{Average token length of responses in model-dataset specific difficulty level. }
    \label{fig: token length by difficulty}
\end{figure}
The number of questions in each category and their average response token length are illustrated in  Table~\ref{tab:question-distribution} and Figure~\ref{fig: token length by difficulty} respectively.
\section{Pair-wise Heatmap of Accuracy, Token Length and Perplexity}\label{app: question level additional analysis}
\begin{figure}[h]
    \centering
    \includegraphics[width=1\linewidth]{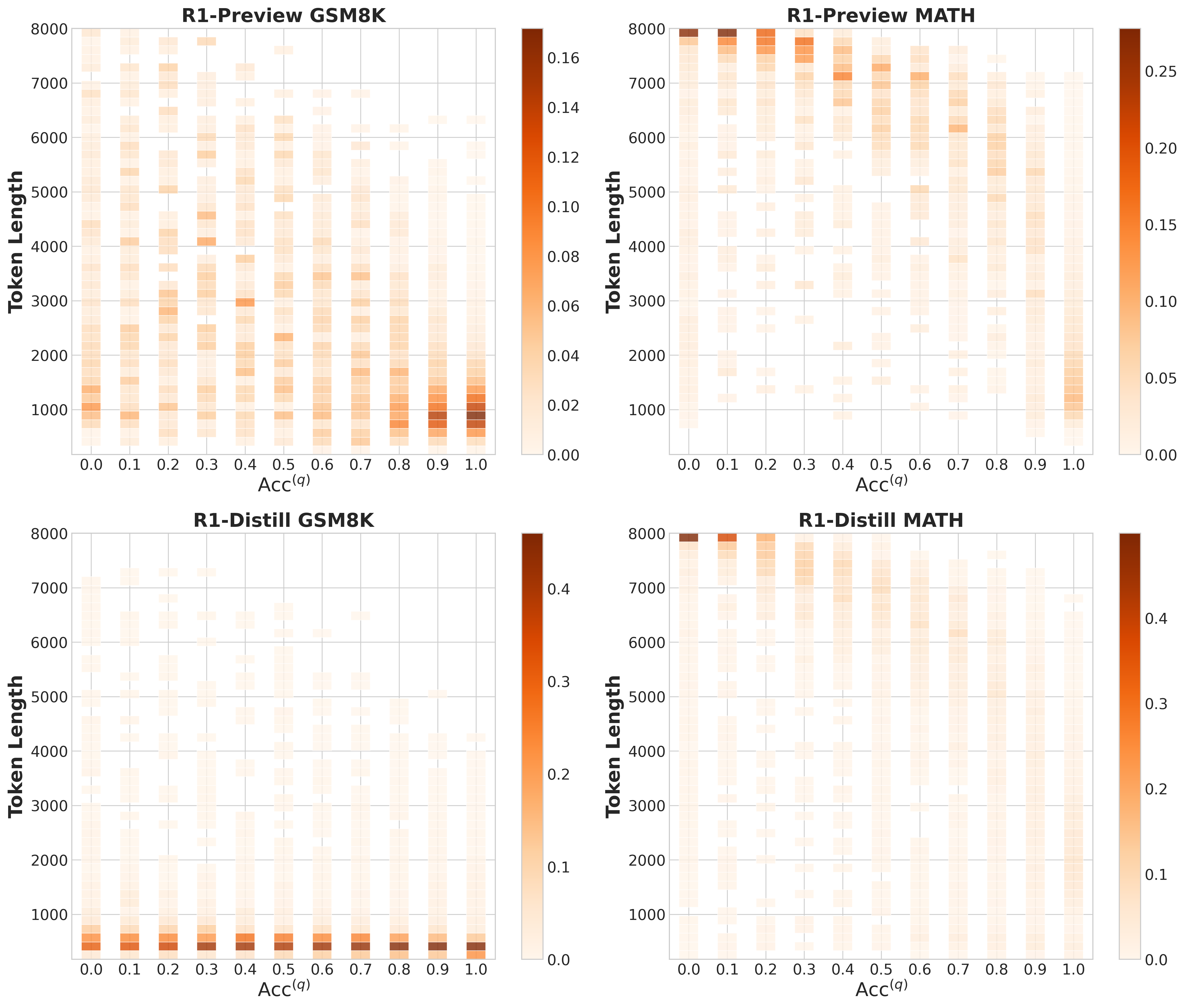}
    \caption{Heatmap of token length for questions categorized by accuracy. Density is normalized among the questions with that accuracy. }
    \label{fig: token len vs acc-question level}
\end{figure}

\paragraph{Length Versus Accuracy} Figure \ref{fig: token len vs acc-question level} shows the heatmap of generation length and accuracy. For easier questions (i.e., questions with high accuracy), the generation lengths tend to cluster around shorter responses.
However, for extremely challenging questions (i.e., questions with accuracy 0), the generation lengths are widely dispersed across the entire range, rather than clustering around the longest response length possible. 

\begin{figure}[h]
    \centering
    \includegraphics[width=1\linewidth]{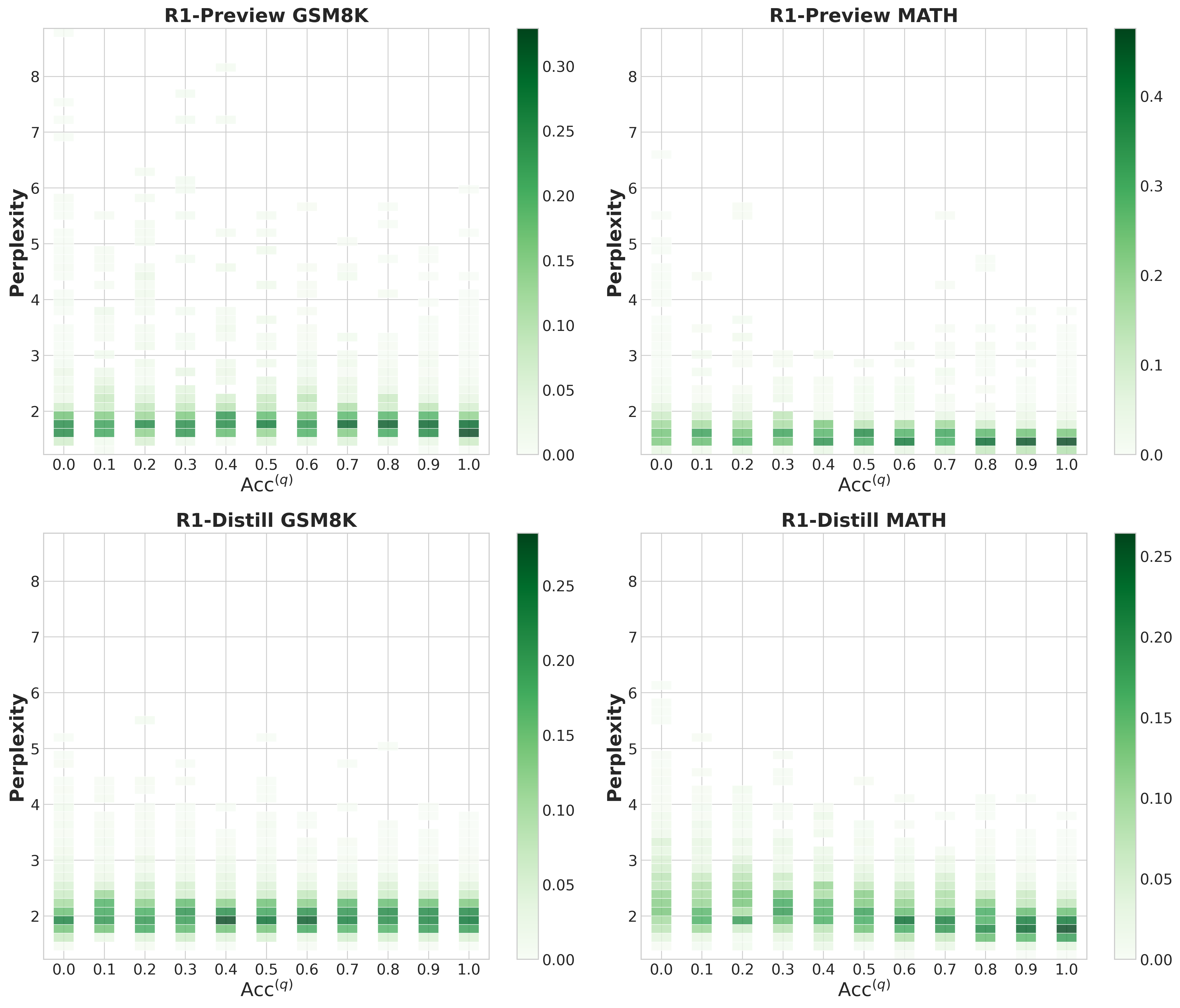}
    \caption{Heatmap of perplexities for questions categorized by accuracy. Density is normalized among the questions with that accuracy.}
    \label{fig: ppl vs acc}
\end{figure}

\begin{figure}[h]
    \centering
    \includegraphics[width=1\linewidth]{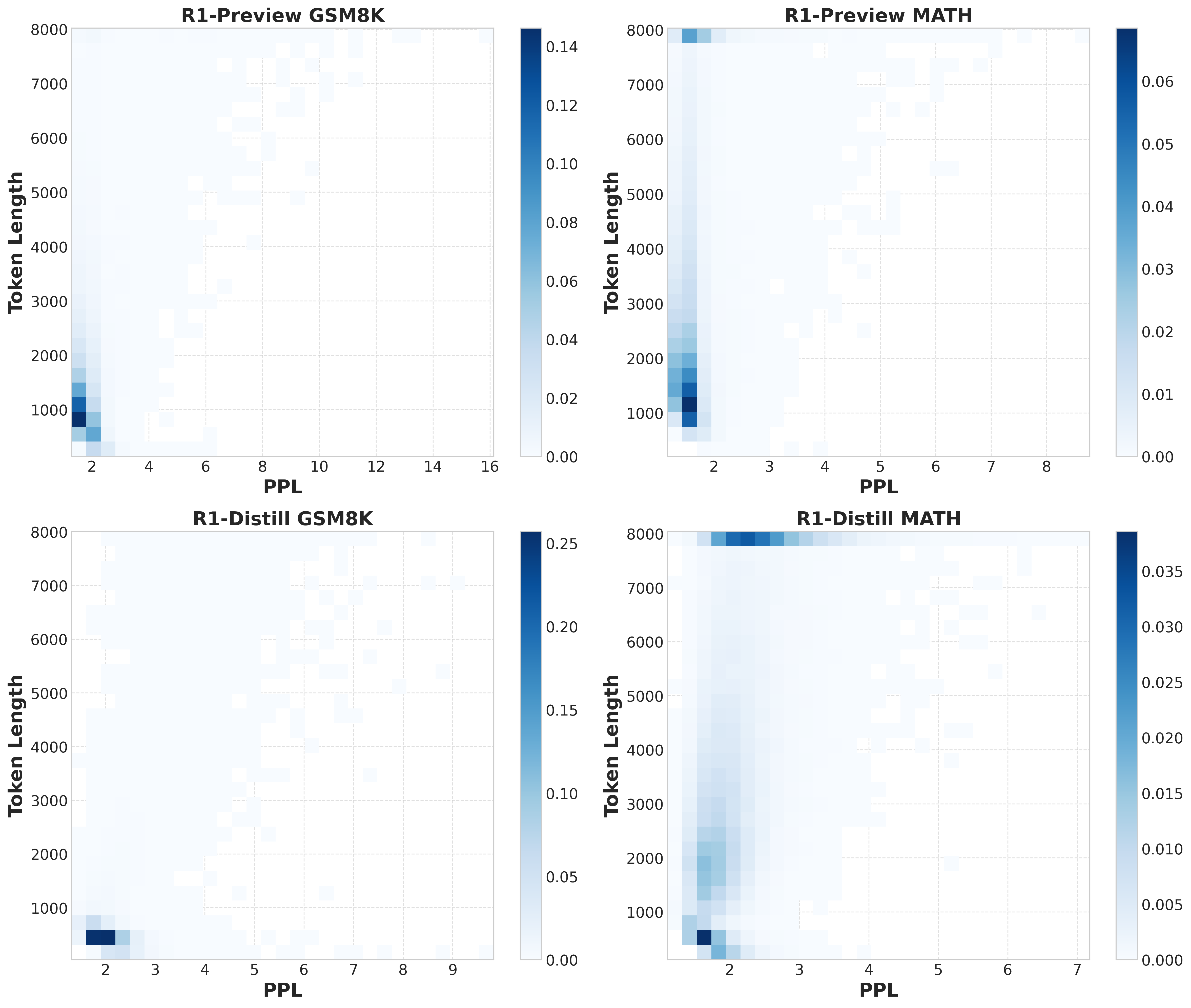}
    \caption{Heatmap of perplexity and token length, the density in the heatmap is normalized on all the questions.}
    \label{fig: ppl vs token len}
\end{figure}

\paragraph{Perplexity Versus Accuracy} Figure \ref{fig: ppl vs acc} shows the heatmap of perplexity and accuracy. Unlike generation length, perplexity doesn’t vary much between easy and hard questions overall, most of the perplexity fall into the same range for questions with various accuracy level. However, we observe for more challenging questions (i.e., those with low accuracy), there are more prominent high-perplexity outliers. In contrast, easier questions exhibit a tighter perplexity distribution with fewer extreme values.
\paragraph{Length Versus Perplexity}
Figure \ref{fig: ppl vs token len} shows the heatmap of generation length versus perplexity. The majority of questions cluster in the region with both low perplexity and short generation length.

\section{Length Reduction using SimPO}\label{app: additional results with simpo}
\begin{figure}[h]
    \centering
    \includegraphics[width=1\linewidth]{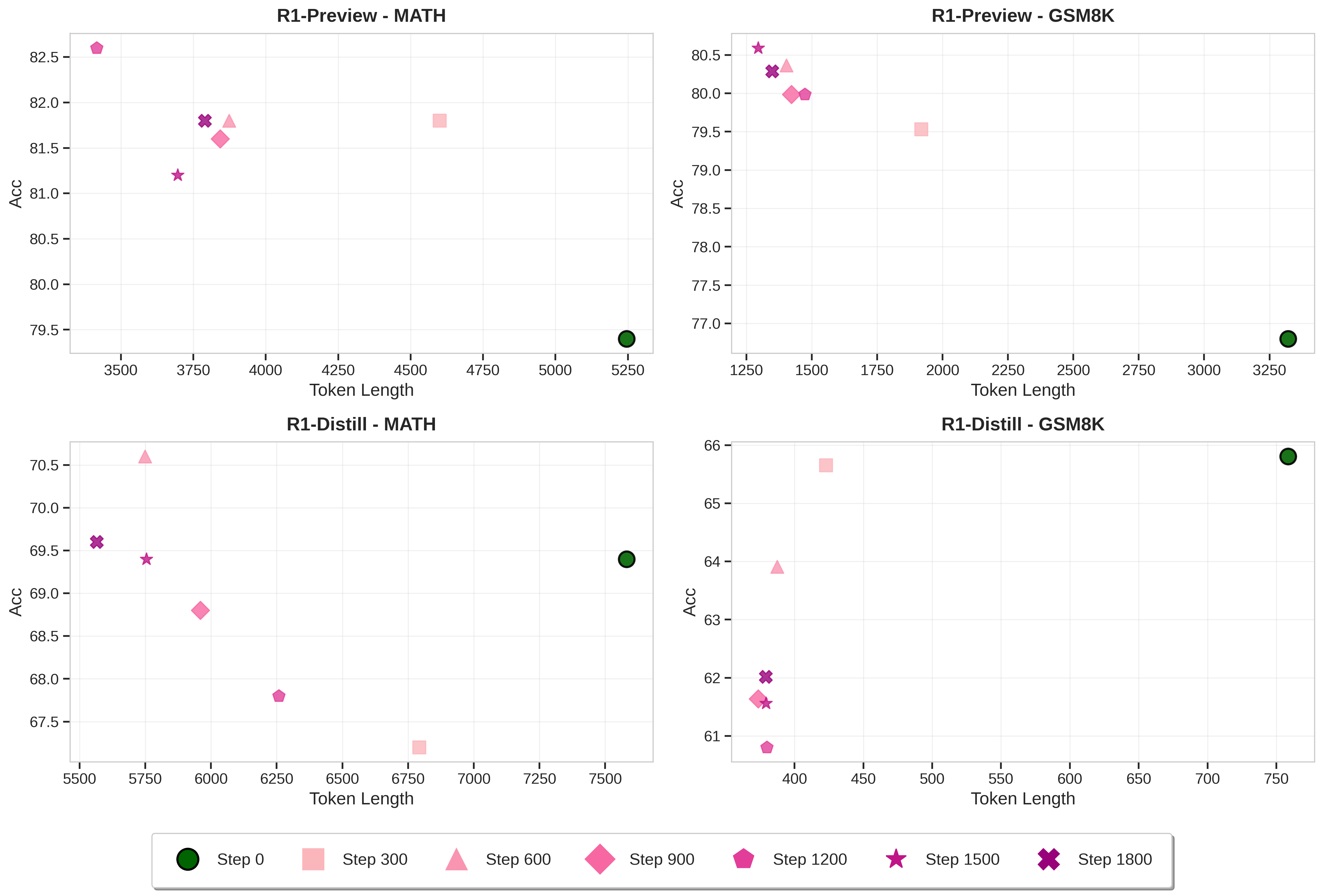}
    \caption{Accuracy and token length trade-offs during SimPO training steps}
    \label{fig: simPO-acc-token scatter}
\end{figure}

\begin{figure}[h]
    \centering
    \includegraphics[width=1\linewidth]{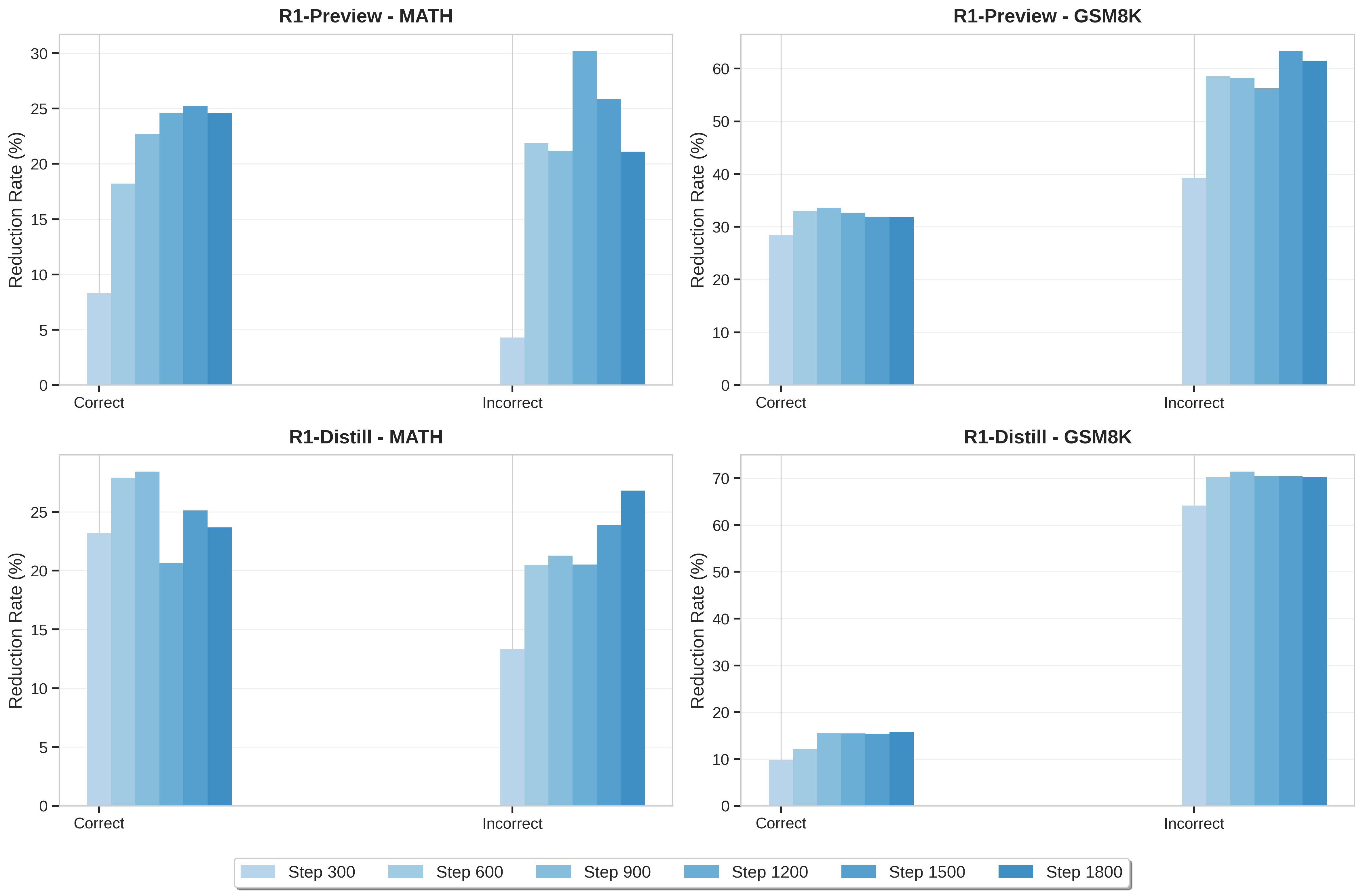}
    \caption{Token length reduction rate for correct and incorrect responses when using SimPO.}
    \label{fig: simPO-token token reduction rate}
\end{figure}

\paragraph{Parameters} In Table \ref{tab:training_hyperparams}, we list the parameters and compute we use when using SimPO to reduce the generation length. 
\paragraph{Additional Results}
In Figure \ref{fig: simPO-acc-token scatter}, we show the accuracy and token length trade-offs during SimPO training steps. In Figure \ref{fig: simPO-token token reduction rate}, we show token length reduction rate for correct answer and incorrect answer during SimPO training steps. For GSM8K dataset, the length reduction on the incorrect responses are significantly high than that of correct responses. While for MATH dataset, the length reduction rate is similar for both correct and incorrect responses. 
\begin{table}[h]
\centering
\caption{Training Hyperparameters using SimPO.}
\begin{tabular}{ll}
\toprule
\textbf{Parameter} & \textbf{Value} \\
\midrule
Learning Rate & \texttt{5e-7} \\
Batch Size (per device) & 1 \\
Gradient Accumulation Steps & 4 \\
SimPO Parameter $\beta$ & 2 \\
SimPO Parameter $\gamma$ & 1 \\
\midrule
GPU & 2 A100 80G \\
\bottomrule
\end{tabular}
\label{tab:training_hyperparams}
\end{table}


\end{document}